%% file: paper.tex
\documentclass[]{bytedance_seed}

\usepackage[toc,page,header]{appendix}

\usepackage{makecell}
\usepackage{graphicx}
\usepackage{subcaption}

\newcommand{\bench}[1]{Workflow-GYM}

\title{Workflow-GYM: Towards Long-Horizon Evaluation  of Computer-use Agentic Tasks in Real-World Professional Fields}

\affiliation[1]{ByteDance Seed}
\affiliation[2]{M-A-P}
\affiliation[3]{Humanlaya}
\affiliation[4]{TokenWave.AI}

\contribution{Full author list in Contributions}

\abstract{
Recent years have witnessed the rapid evolution of AI agents toward handling increasingly complex, real-world tasks. However, existing benchmarks rarely evaluate whether agents can operate graphical user interfaces to complete long-horizon, high-value professional workflows across diverse domains. Current GUI benchmarks still predominantly focus on general-purpose software, relatively simple applications, and short-horizon tasks, leaving it largely unknown whether modern agents can follow user instructions to autonomously operate domain-specific professional software and accomplish economically valuable work in an end-to-end manner. To bridge this gap, we introduce \textbf{Workflow-GYM}, a benchmark for long-horizon GUI tasks centered on professional domains and specialized software environments. Through extensive experiments on state-of-the-art models, we find that even the strongest models achieve only slightly above 30\% success rates, highlighting that professional long-horizon GUI workflows remain highly challenging for current GUI agents. Further analysis reveals that current agents struggle to maintain long-horizon workflow consistency, frequently exhibiting workflow stage omission, error propagation, objective drift, and insufficient understanding of professional software environments. Our findings provide important insights into the limitations of current agent systems and suggest key directions for the next generation of GUI-agent research.

}
\checkdata[Project Page]{\url{https://workflow-gym.github.io/}}

\date{\today}

\usepackage{url}
\begin{document}
\maketitle


\input{sections/introduction}

\input{sections/relatedwork}

\input{sections/approach}
\input{sections/experiments}

\clearpage
\input{sections/contribution}

\clearpage

\bibliographystyle{plainnat}
\bibliography{main}

\clearpage

\beginappendix

\input{sections/appendix}

\end{document}

%% file: sections/introduction.tex
 \section{Introduction}

In recent years, large language models (LLMs) and vision language models(VLMs)  have rapidly evolved beyond text-centric generation and dialogue toward multimodal, agentic systems capable of interacting with external environments\cite{yao2022react,wang2024openhands}. This paradigm shift has reoriented the research focus from solving isolated, short-horizon tasks to enabling models to execute long-horizon, instruction-driven workflows end-to-end. A central question emerging from this transition is whether AI systems can autonomously complete complex tasks that traditionally require sustained human effort and yield tangible economic value. Addressing this question is widely regarded as a critical step toward more general-purpose intelligence.

Within this landscape, GUI Agents occupy a uniquely important position. A large portion of real-world tasks—particularly in professional domains—are performed through graphical user interfaces rather than APIs or command-line tools. Moreover, GUI interactions are inherently observable and interpretable, making them more accessible to non-expert users. In practice, professionals across domains routinely accomplish tasks by operating specialized software through multi-step interactions guided by domain knowledge and task objectives. We abstract such processes as workflows: structured, goal-directed sequences of GUI actions that transform an initial state into a desired outcome. Evaluating whether GUI Agents can successfully execute these workflows provides a principled way to assess their capability to substitute for human labor in real-world settings. Importantly, solving such tasks requires a combination of multimodal perception and action, domain-specific knowledge, tool-use proficiency, and long-horizon reasoning abilities, including planning, memory, and self-correction—thus serving as a comprehensive testbed for advanced agentic intelligence.

However, despite recent progress, existing benchmarks fall short of evaluating this emerging paradigm. Prior GUI-agent benchmarks\cite{liu2025pc,mu2025gui,xie2024osworld,sun2026ambibench} primarily focus on short-horizon, general-purpose tasks involving common applications such as web browsing and basic system operations. As a result, they rarely require interaction with specialized professional software or domain-specific expertise, limiting their ability to reflect realistic, high-value workflows. Meanwhile, benchmarks targeting long-horizon reasoning or economically valuable tasks are largely confined to text-based settings, including code generation, document understanding, and knowledge-intensive QA\cite{patwardhan2025gdpval,ding2025nl2repo,yang2026onemillion}. Although these benchmarks emphasize planning and reasoning, they largely overlook computer-use scenarios that require agents to operate graphical interfaces and interact with real software systems. Consequently, there remains a significant gap: few existing benchmarks systematically evaluate whether agents can execute long-horizon, domain-specialized, economically meaningful workflows through GUI interactions. This gap motivates the need for a benchmark that captures the intersection of professional expertise, real software interaction, and long-horizon agentic execution.

\begin{figure}[t!]
    \centering
    \includegraphics[width=\linewidth]{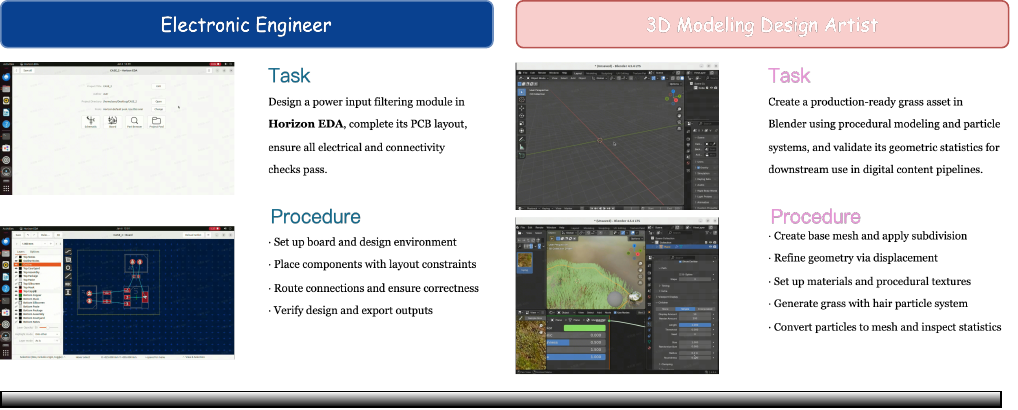}
    \caption{Examples of Workflow-GYM tasks from professional domains. Each task requires interacting with specialized software through graphical user interfaces to accomplish a real-world objective.}
    \label{fig:demo}
\end{figure}

To bridge this gap, we introduce \textbf{Workflow-GYM}, a benchmark designed to evaluate whether GUI Agents can execute end-to-end, domain-specific workflows in realistic working environments. In Workflow-GYM, each task is instantiated in a fully configured environment with the required professional software pre-installed, and the agent is provided only with a natural language instruction describing the task objective—without access to intermediate guidance or additional input hints. Starting from this minimal specification, the agent must autonomously plan, reason, explore, and interact with the environment to progressively achieve the desired goal.
Evaluation is performed based on the final state or artifact produced by the agent. For each task, we define a unique and automatically verifiable success criterion, enabling deterministic assessment of task completion. This design ensures both evaluation reliability and scalability, while faithfully reflecting whether an agent can successfully carry out the entire workflow and produce the intended outcome in a real-world setting. Typical tasks of Workflow-GYM is shown at Figure~\ref{fig:demo}.

Through extensive experiments with multiple state-of-the-art(SOTA) models under different agent frameworks, we find that long-horizon, domain-specific software workflows remain highly challenging for current systems. 
Even the strongest model can only achieve a success rate of 30\%, 
highlighting a substantial gap between existing capabilities and the requirements of real-world tasks.
Further analysis reveals several recurring failure modes, including workflow stage omission, error propagation, objective drift, and insufficient software-specific knowledge. Beyond these model-level limitations, Workflow-GYM also exposes a more fundamental challenge shared by current GUI-agent frameworks: the mismatch between continuous human interaction and the discrete observation-action paradigm adopted by most existing agents.

To summarize, our contributions are as follows:
\begin{itemize}
  \item We introduce the concept of \textit{workflow} as a unified abstraction of real-world, long-horizon, economically meaningful compute-use processes performed by domain experts, and construct \textbf{Workflow-Gym}, a benchmark that captures such tasks across diverse professional domains.

  \item We develop a reliable and reusable data construction pipeline, combining expert-driven design, multi-stage validation, and environment-grounded execution, enabling scalable generation of high-quality, verifiable workflow tasks.

  \item We conduct a comprehensive evaluation of representative GUI agents and foundation models on \textsc{Workflow-Gym}, revealing substantial performance gaps and systematically analyzing key challenges, including long-horizon level failures, insufficient professional software knowledge, and repetitive action looping.
\end{itemize}

%% file: sections/relatedwork.tex
\section{Related Works}
\subsection{Benchmarks for GUI Agents and Long-Horizon tasks}

Existing GUI benchmarks can be characterized along two complementary dimensions:  interaction device and task scenario. From the device perspective, they include (1) PC-based benchmarks\cite{xie2024osworld,dai2025scuba} and (2) mobile benchmarks\cite{rawles2024androidworld}, which differ in interface structure and interaction paradigms. From the task perspective, they can be grouped into (a) general-purpose benchmarks that focus on daily common UI operations, and (b) domain-specific(or software-specific) benchmarks that target specialized applications requiring domain knowledge. For instance, AmbiBench\cite{sun2026ambibench} focuses on everyday user activities in mobile environments, such as app-based services like food ordering, reflecting general-purpose interaction scenarios, while GUI-360\cite{mu2025gui} evaluates agents’ ability to operate productivity software such as Microsoft Office, and Science-Board\cite{sun2025scienceboard} targets scientific workflows in research environments, representing domain-specific or software-specific settings.
Despite this diversity, existing benchmarks are largely limited to short-horizon interactions and fail to capture end-to-end, long-horizon workflows that arise in real-world professional settings across diverse domains, particularly those requiring the use of domain-specific software and complex multi-stage execution.

\subsection{MLLM-Driven GUI Agents}
Compared to GUI-related question answering benchmarks\cite{shi2025gui}, end-to-end GUI interaction tasks more effectively reflect a model’s ability to interact with real graphical user interfaces and accomplish practical objectives.
The rapid advancement of LLMs and VLLMs has driven the emergence of GUI agents capable of interacting with GUI environments to accomplish tasks.Mobile-Agent-v2\cite{deng2024mobile} studies GUI interaction in mobile environments and formulates it as a multi-agent collaboration problem, introducing specialized planning, decision, and reflection agents to handle long-horizon execution over interleaved visual-text contexts. There are also researches not only proposing agent frameworks but also training dedicated models to better adapt to the target environments and interaction dynamics.
UI-TARS\cite{wang2025ui} develops a native agentic framework that unifies reasoning–action–observation loops with a sandboxed execution environment and scalable rollout infrastructure, enabling end-to-end training of GUI agents via multi-turn reinforcement learning. Step-GUI\cite{yan2025stepguitechnicalreport} proposes a GUI agent built on a step-wise interaction framework with calibrated step-level rewards (CSRS), trained via a self-evolving pipeline that iteratively improves action grounding and long-horizon task execution from large-scale GUI trajectories.

%% file: sections/approach.tex
\section{Workflow-GYM}

\subsection{Dataset Construction}

To faithfully evaluate the capability of GUI agents in completing real-world, high-value tasks, the benchmark must capture several key properties of professional workflows: Tasks should be grounded in authentic real-world scenarios to prevent retrieval-based solutions, require domain-specific knowledge and professional software, exhibit long-horizon multi-step complexity, and admit clear, objective success criteria for reliable and reproducible evaluation.
Guided by these requirements, we design a multi-stage data construction pipeline that systematically sources, filters, instantiates, and validates tasks, ensuring both realism and evaluation rigor.

\begin{figure}[t!]
    \centering
    \includegraphics[width=\linewidth]{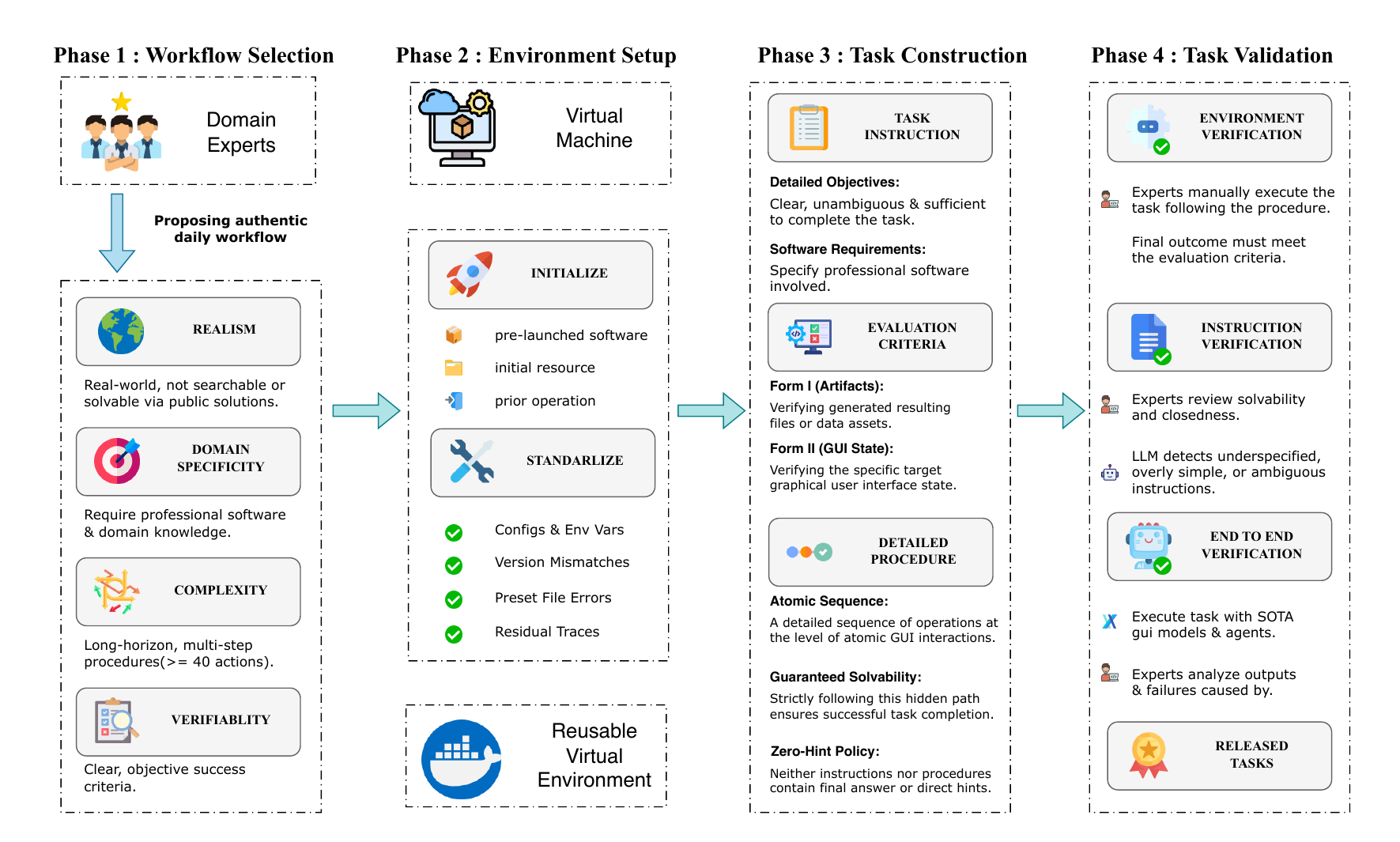}
    \caption{Construction pipeline of Workflow-Gym tasks.}
    \label{fig:frame}
\end{figure}

\subsubsection{Workflow Selection}

We begin the construction pipeline by sourcing candidate tasks from domain experts across multiple professional fields. To ensure that the resulting benchmark reflects realistic and practically meaningful workflows, experts are instructed to propose tasks grounded in their own daily work practices rather than synthetic or publicly documented examples. The background of expert team is shown at Appendix~\ref{app:annotator}.

To further enforce the desired task properties, we apply a set of strict filtering criteria. Specifically, each task must satisfy all of the following requirements:

\begin{itemize}
    \item \textbf{Realism}. Tasks must originate from authentic real-world workflows and should not be directly searchable or solvable via publicly available solutions, ensuring that success cannot be achieved through simple retrieval or memorization.
    \item \textbf{Domain Specificity}. Completing the task must require the use of professional software and domain-specific knowledge, rather than general-purpose tools alone.
    
    \item \textbf{Complexity}. Tasks must involve long-horizon, multi-step procedures, with a minimum of 30 atomic actions, to reflect realistic workflow complexity.
    
    \item \textbf{Verifiability}. Each task must admit a clear and objective success criterion, enabling a reliable evaluation of whether the task has been completed correctly.
    
\end{itemize}

Applying these criteria, we curate an initial pool of over 1,000 candidate tasks that serve as the foundation for subsequent stages of dataset construction.

\subsubsection{Environment Setup}

To ensure consistency, reproducibility, and fair comparison across tasks, we adopt a two-level environment construction strategy that separates reusable software environments from task-specific instantiation.

\textbf{Base Environments: reusable, software-level execution backbones.}
Instead of constructing a fully isolated environment for each individual task, we build standardized base environments at the level of professional software and its corresponding version. Each base environment is packaged as a standalone virtual machine image with all required software pre-installed and configured. This design significantly improves scalability and reuse, while preserving strict control over software dependencies. Compared to containerized or API-based setups, full-system virtual machines more faithfully capture real-world GUI interactions, including rendering, window management, and system-level dependencies. In this phase we construct 56 independent virtual machines, each representing the execution workspace for a specific professional software. The full list of softwares  in \bench{} is shown at Appendix~\ref{app:softwarelist}.

\textbf{Task Instantiation: lightweight, task-specific environment initialization.}
To account for task-specific requirements, we perform a lightweight instantiation step at runtime. For each task, necessary artifacts—such as input data, templates, or configuration files—are injected into the workspace through predefined initialization procedures. All environments are initialized to a unified interaction-ready state: required software is pre-launched, but no task-specific operations have been performed. Irrelevant preconditions, such as login or authentication steps, are removed to eliminate extraneous interactions.

This modular design mirrors real-world workflows, where multiple tasks are executed within a shared software environment but differ in their inputs and objectives. As a result, we isolate agent behavior as the sole source of variation while enabling strictly controlled, repeatable, and directly comparable evaluation across tasks and agents.

\subsubsection{Task Construction}

Based on the filtered workflow candidates and environments built in previous stages, we construct each task on a per-case basis by transforming high-level workflows into fully specified and executable benchmark instances. Together with the environment, each task is composed of the following components:

(1) \textbf{Task Instruction.}
Domain experts provide a complete and self-contained task description, including necessary background, detailed objectives(final state, artifacts and directories), and the professional software involved. The instruction is required to be clear, unambiguous, and sufficient for completing the task without relying on external information.

(2) \textbf{Expected Outcome and Evaluation Criteria.}
Each task is associated with a well-defined target outcome and corresponding evaluation criteria. Depending on the task type, the outcome takes one of two forms: (i) a resulting artifact (e.g., files or documents), or (ii) a target graphical user interface (GUI) state. In both cases, experts specify explicit requirements that the final output must satisfy. Evaluation is expected to perform in a binary manner, where a task is considered successful if and only if all criteria are met.
To ensure robustness and reproducibility, the evaluation criteria are restricted to deterministic and reliably reproducible aspects of the outcome. Factors that are inherently stochastic or environment-dependent are excluded from evaluation whenever possible.

(3) \textbf{Step-by-Step Expert Procedure.}
To enhance verifiability and ensure task solvability, experts provide a detailed sequence of operations required to complete the task, specified at the level of atomic GUI interactions. These procedures are constructed such that strictly following them leads to successful task completion under the defined evaluation criteria. Importantly, neither the task instructions nor the procedural annotations contain the final answer or any direct hints that would trivially reveal it.

\subsubsection{Task Validation}
To ensure the quality and reliability of tasks, we adopt a multi-stage validation mechanism:
\begin{itemize}

\item \textbf{Environment verification}:
To ensure that task failure is not caused by environmental factors, domain experts manually execute each task within the designated environment by strictly following the provided procedural instructions. The final outcome must satisfy the predefined evaluation criteria. If discrepancies arise, targeted revisions are performed depending on the root cause, including adjustments to the environment setup, evaluation criteria, or procedural instructions.

 \item \textbf{Instruction-level validation}:
We conduct a dual-layer quality check on task instructions, combining expert review with LLM-as-a-judge evaluation. Senior domain experts assess whether task descriptions satisfy key properties such as closedness (i.e., all necessary information is provided) and solvability. In parallel, the LLM-based evaluator identifies issues including underspecified instructions, excessive simplicity (e.g., insufficient procedural depth), and ambiguous operation descriptions that prevent unambiguous execution. Tasks failing to meet these criteria are returned to annotators for revision.

\item \textbf{End-to-end preliminary experiment refinement}:
We further validate task quality through end-to-end agent-based experiments. Specifically, task instructions are executed by a SOTA GUI agent within the target environment, and both execution results and trajectories are collected. Human experts then analyze these outputs to identify potential issues unrelated to model capability, such as latent ambiguities in instructions, missing constraints, or environment inconsistencies (e.g., software version or configuration mismatches). Identified issues trigger corresponding refinements to the task specification or environment.
\end{itemize}

Only tasks that successfully pass all three validation stages are retained in the final benchmark. Tasks failing at any stage are iteratively revised until all validation criteria are satisfied.

\subsection{Benchmark Statistics}
\begin{figure}[t!]
    \centering
    \includegraphics[width=0.7\linewidth]{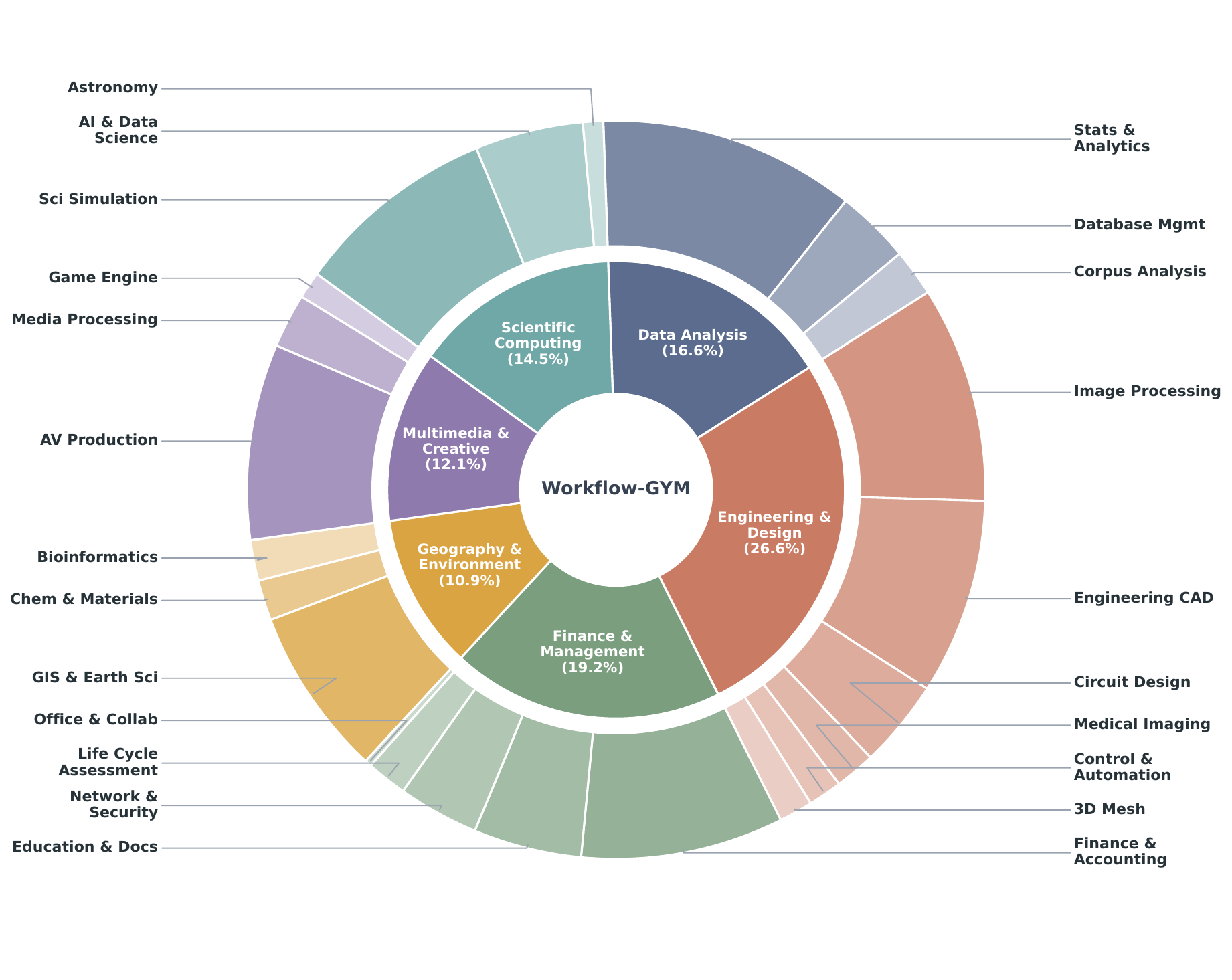}
    \caption{Task domains and distributions of \bench~. }
    \label{fig:distribution}
\end{figure}

Finally we construct \textbf{Workflow-GYM}, a benchmark evaluating the capabilities of GUI agents on completing real-world professional computer-use tasks in different domains and generate revenue.   
Workflow-GYM comprises a total of \textbf{338} tasks, spanning 6 top-level domains and 23 fine-grained subdomains, ensuring broad coverage across diverse application areas. Detailed domain distribution is shown in table \ref{fig:distribution}. Beyond domain diversity, the tasks exhibit substantial variation in procedural complexity: the number of execution steps required to complete a task ranges from 30 to 110. Based on this variation, we categorize task difficulty according to the length of the execution procedure shown in table \ref{tab:difficultyofwg}, using the number of required steps as a proxy for complexity.

\begin{table}[t]
\centering
\caption{Task difficulty distribution in Workflow-GYM based on the number of execution steps.}

\begin{tabular}{lccc}
\toprule
Difficulty Level & Step Range & Task Number & Percentage \\
\midrule
Easy & 30--44 & 129 & 38.2\% \\
Medium & 45--60 & 159 & 47.0\% \\
Hard & 61--110 & 50 & 14.8\% \\
\bottomrule
\end{tabular}
\label{tab:difficultyofwg}
\end{table}

\subsection{Automatic Evaluation for Workflow-GYM Tasks}
For each task in Workflow-GYM, we conduct automated evaluation based on its final outcome, either in the form of a produced artifact or an ultimate system state. For tasks without explicit output artifacts (i.e., non-artifact tasks), we capture \textbf{screenshots of the GUI} after task completion and, together with the predefined evaluation criteria, provide them to a vision-language model (VLM) for scoring. For tasks that yield concrete artifacts, we assess correctness by verifying whether the generated files satisfy the specified requirements. Depending on the characteristics of each task, we adopt either rule-based or LLM-based evaluation methods to ensure reliable and scalable assessment.


%% file: sections/experiments.tex
\section{Experiments}
\subsection{Experimental Settings}

We evaluate a set of representative frontier models on Workflow-GYM, including GPT-5.4-xhigh, GPT-5.4-mini-xhigh~\cite{openai2026gpt5}, Gemini-3.1-Pro, Gemini-3-Flash~\cite{GoogleCloud2026Gemini3.1Pro,GoogleCloud2026Gemini3flash}, Kimi-k2.6~\cite{moonshot2026k2_6} and Seed-2.0-Lite \cite{seed2026seed2}. In our setting, each model receives a single natural-language instruction at the beginning of the task and must complete the entire workflow through GUI interactions within an isolated environment, without any intermediate guidance or additional prompts. Given that computer-use capabilities are tightly coupled with model-specific architectures, training paradigms, and tool-use interfaces, we adopt model-specific agent frameworks to ensure effective interaction but set the maximum of interaction rounds to 400 altogether\footnote{The process which agent receives a screenshot images and then outputs actions are calculated as one round. }. To account for the inherent stochasticity and variability in GUI-based task execution, we run each task for three independent trials (trialnum = 3) and report aggregated results. During evaluation, for all non-rule-based scoring cases, we employ Seed-1.8\cite{seed2026seed1} as the judge model. Since the judge operates strictly based on predefined and validated scoring rubrics, its influence on evaluation bias is minimal.

\subsection{Main Results}

\begin{table}[t]
\centering
\small
\begin{tabular}{lcc|ccc}
\toprule
\textbf{Model} & \textbf{Avg Pass (\%)} & \textbf{Pass@3} & \textbf{Easy} & \textbf{Medium} & \textbf{Hard}   \\
\midrule
Gemini-3.1-pro  & \textbf{30.67} & 41.12 & 39.28 & 26.62 & 21.33   \\
Kimi-k2.6     & 29.68 & \textbf{41.42} & 34.63 & 27.04 & 25.33  \\
GPT-5.4         & 17.85 & 26.33 & 21.71 & 16.14 & 13.33   \\
Seed-2.0-lite   & 18.24 & 28.40 & 24.91 & 15.09 & 11.33 \\
GPT-5.4-mini    & 15.98 & 27.22 & 18.09 & 15.51 & 12.00  \\
Gemini-3-flash  & 7.89  & 15.98 & 12.14 & 6.08  & 2.67   \\
\bottomrule
\end{tabular}
\caption{Updated performance of different models on Workflow-GYM. Avg Pass denotes overall success rate. Results are averaged over $338*3=1014$ trials.}
\label{tab:workflow_gym_updated}
\end{table}
\begin{figure}[t!]
    \centering
    \includegraphics[width=0.75\linewidth]{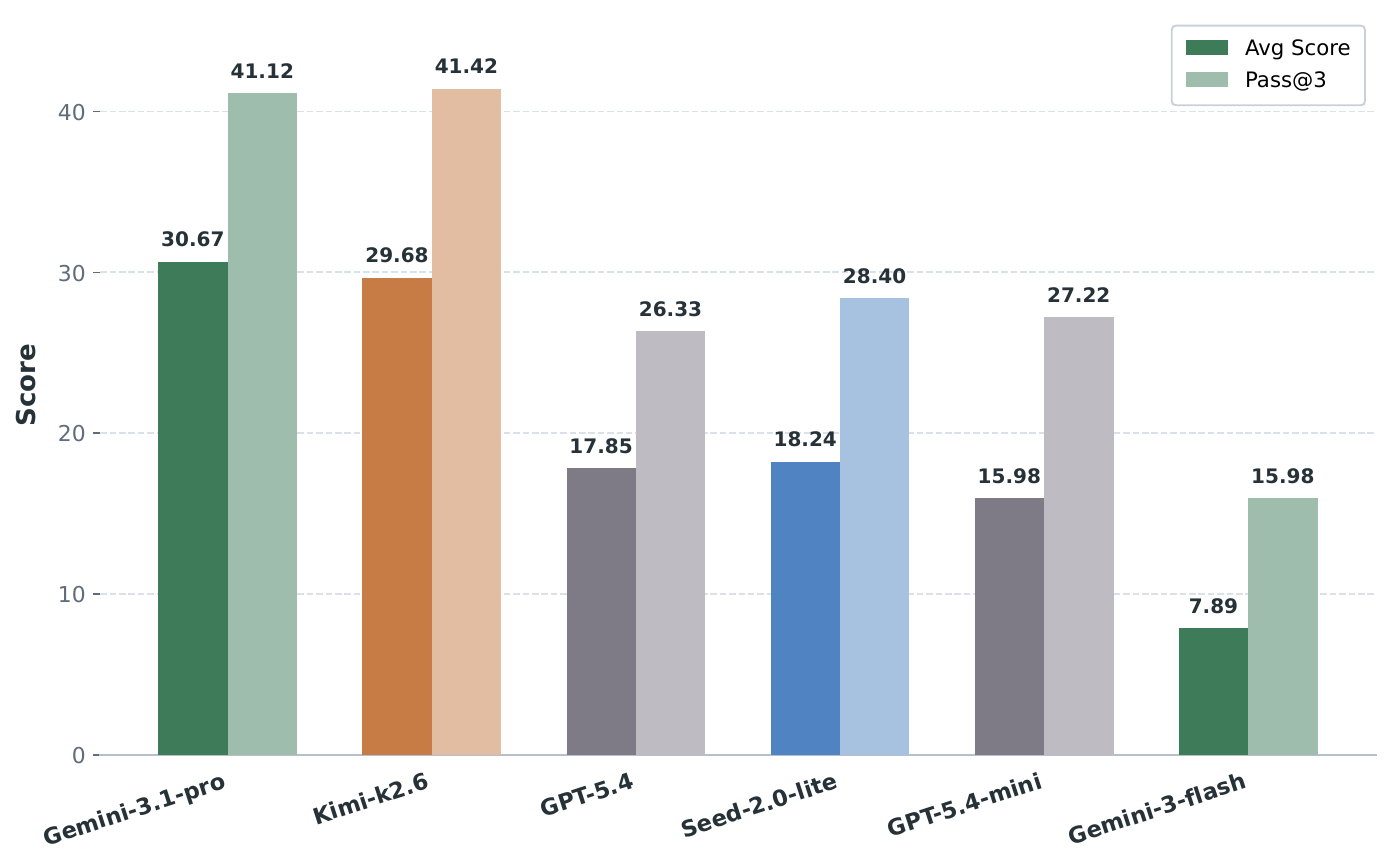}
    \caption{Leaderboard of Workflow-GYM.}
    \label{fig:workflow_gym_leaderboard}
\end{figure}

As shown at the results in Table~\ref{tab:workflow_gym_updated}, several significant observations emerges:

\textbf{Long-Horizon professional task is still a challenging field for GUI-Agents and models.} As shown in Table~\ref{tab:workflow_gym_updated}, even the best-performing models, Gemini-3.1-pro and Kimi-k2.6, merely achieves an average pass rate of 30.67\%  and pass@3 rate of 41.42\%, respectively. Moreover, the majority of models exhibit pass@3 rates below 30\%. In contrast, SOTA models report success rates exceeding 70\% on benchmarks such as OS-World\cite{xie2024osworld}. This substantial performance gap suggests that, while current models and agent frameworks demonstrate strong capabilities in short-horizon tasks and interactions with commonly used general-purpose software, they remain significantly limited in professional, domain-specific settings. In particular, their ability to autonomously execute long-horizon, multi-step workflows in specialized software environments and complete tasks end-to-end at a human level still requires considerable improvement.

\textbf{Model Performance Degrades severely when Step Number Increases}

As shown in Table~\ref{tab:workflow_gym_updated}, all models exhibit a clear performance degradation as task difficulty increases. This trend suggests that long-horizon workflow tasks pose substantially greater challenges for current GUI agents.
Unlike short atomic interactions, completing a realistic workflow requires the agent to perform a sequence of interdependent decisions and executions in the correct order. Failure at any intermediate step can propagate through subsequent stages and ultimately lead to overall task failure. As workflow length increases, the difficulty of maintaining coherent planning and reasoning across the entire trajectory also grows significantly.

Moreover, longer workflows inherently increase the likelihood of encountering complex low-level interactions, such as precise dragging, multi-step UI manipulations, or coordination between keyboard and mouse operations. These operations are particularly error-prone for current agents, further amplifying the probability of failure in long-horizon tasks. Collectively, these results indicate that both long-range sequential reasoning and reliable execution of complex interactions remain major bottlenecks for existing GUI agents.

\textbf{Domain Performance Difference}
\begin{table*}[t]
\centering
\small
\resizebox{\textwidth}{!}{
\begin{tabular}{lcccccc}
\toprule
\textbf{Model} & \textbf{Data Analysis} & \textbf{Eng \& Design} & \textbf{Fin \& Management} & \textbf{Geo \& Environment} & \textbf{MM \& Creative} & \textbf{Sci Computing} \\
\midrule

Gemini-3.1-pro  & 39.88 & \textbf{24.44} & \textbf{37.44}& \textbf{27.93} & 21.95 & \textbf{31.97} \\
Kimi-k2.6& \textbf{45.24} & 21.48 & 32.82 & 25.23 & \textbf{27.64} & 27.89 \\
GPT-5.4         & 26.19 & 10.37 & 24.10 & 17.12 & 17.89 & 14.29 \\
GPT-5.4-mini    & 19.05 & 12.59 & 20.00 & 9.01  & 17.07 & 17.69 \\
Seed-2.0-lite   & 16.07& 10.74 & 24.62& 18.02 & 18.70 & 25.85\\
Gemini-3-flash  & 10.12 & 6.67  & 9.23  & 0.90  & 8.94  & 10.20\\
\midrule
All Models Avg. & 26.09& 14.38& 24.70& 16.37& 18.08& 21.32\\
\bottomrule
\end{tabular}
}
\caption{Performance breakdown across different primary domains in Workflow-GYM. Results are reported as average pass rates (\%).}
\label{tab:domain_result}
\end{table*}

Table~\ref{tab:domain_result} reveals substantial performance variation across different workflow domains, suggesting that current GUI agents exhibit distinct capability biases. Kimi-k2.6 achieves the strongest performance on \textit{Data Analysis} and \textit{Multimedia \& Creative} tasks, likely benefiting from its strong coding and multimodal perception  abilities. Gemini-3.1-pro performs particularly well on \textit{Geography \& Environment}, \textit{Engineering \& Design}, and \textit{Scientific Computing} tasks, which may be related to its stronger scientific  and reasoning capabilities.

More broadly, current agents perform significantly better on workflows with relatively structured and standardized GUIs, such as data analysis or finance-related tasks, while struggling on more visually open-ended environments such as multimedia creation. One possible reason is that structured GUIs typically contain semantically explicit interface elements, stable layouts, and relatively discrete interactions, making them easier for agents to parse and execute reliably. In contrast, open-ended graphical environments often require continuous spatial manipulation, stronger visual grounding, and fine-grained control over complex interfaces, posing substantially greater challenges for existing GUI agents.

\textbf{Agentic framework matters.}  While many models can achieve strong performance under a unified, general-purpose framework in coding scenarios\cite{wang2024openhands,merrill2026terminal}, GUI-based tasks exhibit a high degree of model–framework coupling. In particular, different models often require tailored configurations of the agent loop, tool group design, and pixel-space grounding mechanisms to perform effectively.
This discrepancy primarily arises from the fact that models are typically trained under specific GUI environment assumptions and tool schemas. For instance, certain models such as Gemini produce spatial coordinates in a normalized [0,1000] grid rather than the native pixel space of the interface. As a result, their action outputs implicitly rely on a corresponding transformation layer within the agent framework.
Consequently, once a model is deployed outside of its aligned agent loop and configuration—without appropriate adaptation of coordinate systems, action representations, or tool interfaces—its performance degrades significantly. This highlights the lack of a universally compatible GUI agent framework and underscores the importance of model-specific adaptation for reliable evaluation and deployment.

\section{Analysis and Discussions}

\subsection{Outcome-level Failure Decomposition}

To better understand the causes of failure on Workflow-GYM from an outcome perspective, we decompose unsuccessful executions into two categories that naturally arise from the long-horizon and end-to-end nature of workflow tasks.

\begin{itemize}
    \item \textbf{Workflow Incompletion }refers to cases where the agent fails to complete the entire workflow and never reaches the required final state. Typical examples include failing to generate the required output artifact, terminating before completing all necessary operations, or leaving the software in an intermediate state rather than the target state specified by the task. 
    
    \item \textbf{Final-State but Incorrect} conversely refers to cases where the agent successfully reaches the final stage of the workflow and produces the expected artifact or GUI state, but the delivered result does not satisfy the task requirements due to parameter mismatches, missing components, incorrect configurations, or other execution errors.

\end{itemize}

\begin{figure}[t]
    \centering
    \begin{subfigure}[b]{0.48\linewidth}
        \centering
        \includegraphics[width=\linewidth]{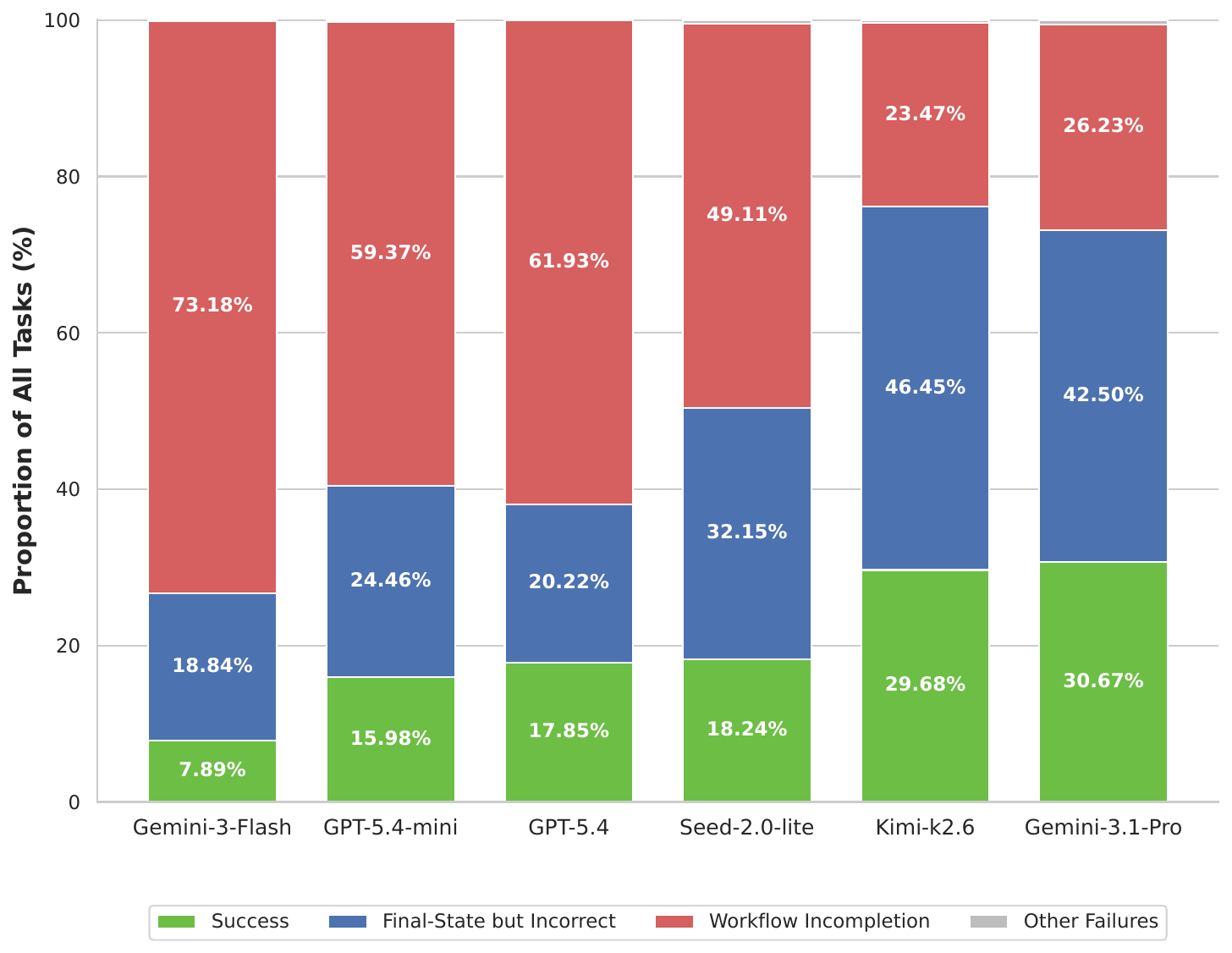}
        \caption{Proportion of different outcome-level failure reason of different models on Workflow-GYM.}
        \label{fig:a}
    \end{subfigure}
    \hfill
    \begin{subfigure}[b]{0.48\linewidth}
        \centering
        \includegraphics[width=\linewidth]{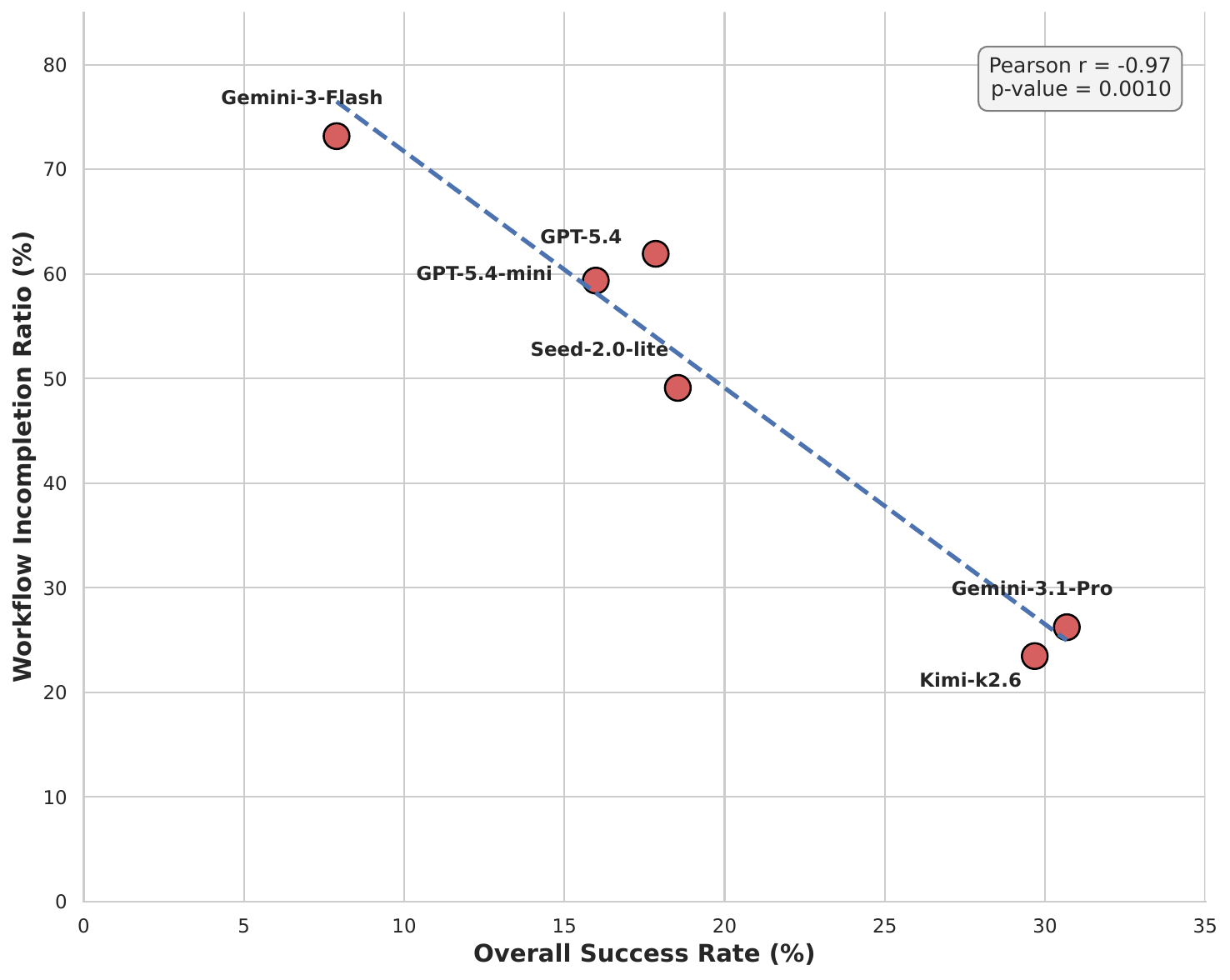}
        \caption{Correlation between model performances and workflow-incompletion rates.}
        \label{fig:b}
    \end{subfigure}

    \caption{Outcome-Level failure calculation of Workflow-GYM.}
    \label{fig:main}
\end{figure}

Figure \ref{fig:a} presents the distribution of these two failure categories across different models. A clear trend emerges: models with higher overall success rates exhibit substantially lower workflow incompletion rates, whereas lower-performing models fail predominantly because they are unable to complete the workflow. To further quantify this relationship, we compute the correlation between workflow incompletion rate and overall model performance. As shown in Figure~\ref{fig:b}, the two variables exhibit a strong negative correlation (Pearson $r=-0.97$), indicating that workflow incompletion decreases consistently as model capability improves.

This observation suggests that, for current models and GUI agents, successfully completing a long-horizon workflow remains a primary bottleneck. In particular, weaker models often fail before reaching the final stage of the task, implying deficiencies in long-range planning, state tracking, error recovery, and sustained execution. By contrast, stronger models are considerably more likely to complete the workflow, and their failures increasingly arise from execution-level inaccuracies after reaching the final stage. In other words, as model capability improves, the dominant failure mode gradually shifts from \textbf{failing to finish the workflow} to \textbf{finishing the workflow but producing an incorrect result}. This transition highlights that workflow completion itself is a critical capability barrier for current computer-use agents and underscores the unique challenges posed by long-horizon professional workflows.

\subsection{Behavior-level Failure Analysis}

While outcome-level analysis characterizes the observable consequences of unsuccessful executions, it does not reveal the underlying mechanisms that lead to failure. To better understand the challenges posed by long-horizon professional workflows, we conduct a trajectory-level analysis and identify two dominant categories of process-level failures: \textbf{Long-Horizon-Level Failures} and \textbf{Execution-Level Failures}.

\textbf{Long-Horizon-Level Failures} arise from the difficulty of maintaining coherent task execution over extended sequences of interdependent actions, including failures in cross-step consistency, progress tracking, and completion verification. In contrast, \textbf{Execution-Level Failures} originate from deficiencies in interacting with professional software environments, including inaccurate visual grounding, insufficient software-specific knowledge, and failures in complex GUI manipulation. We analyze these two categories separately in the following sections.

\subsubsection{Long-Horizon-Level Failures}

\textbf{Error Propagation:} The agent performs an incorrect operation that introduces a deviation from the intended workflow state. Since professional workflows typically consist of tightly coupled stages with strong inter-stage dependencies, subsequent actions often rely on the correctness of preceding states. Once execution diverges from the intended trajectory, later actions may remain locally plausible but are performed under invalid assumptions and therefore fail to make meaningful progress toward the task objective. Without effective recovery, the deviation persists and propagates through subsequent stages, ultimately leading to task failure.

Figure \ref{fig:blender} illustrates a representative example from \textit{Blender}. The agent incorrectly selects a \textit{Plane} instead of the required \textit{Cylinder} at an early stage, causing the execution trajectory to diverge from the intended workflow.Due to the agent's incorrect early selection which went undetected, the system entered an unintended state. All subsequent workflows and operations were built upon this deviated state, rendering all efforts futile.
Although subsequent actions remain locally plausible, they are performed on an invalid workflow state and ultimately fail to achieve the task objective.

\textbf{Workflow Stage Omission:} The agent skips a mandatory intermediate stage in a multi-stage workflow and proceeds directly to subsequent operations. Because many professional workflows contain strong stage dependencies, later actions often rely on artifacts or states produced by earlier stages. As a result, the execution trajectory diverges from the intended workflow, and the task may become unrecoverable even though the agent continues making seemingly reasonable progress. This failure mode is particularly prevalent in long-horizon workflows, where errors introduced at one stage can propagate across many subsequent steps. A typical example is shown at Appendix \ref{app:omission case}.

\textbf{Objective Drift:} The agent gradually deviates from the original workflow objective and becomes overly focused on intermediate interface states or local subgoals encountered during execution. In long-horizon professional workflows, many actions only serve as transitional steps toward a final outcome. However, once the agent mistakenly treats an intermediate state as the objective itself, subsequent decisions become guided by this incorrect target. As a result, the execution trajectory may remain locally consistent and operationally plausible, while progressively diverging from the intended workflow and ultimately failing to satisfy the original task requirement. Example of objective drift in \bench{} is shown at Appendix \ref{app:objdrift case}.

\subsubsection{Execution-Level Failures}

Prior work on GUI agents has documented common execution-level errors, such as mis-clicks or misinterpreting interface elements. Beyond these, Workflow-GYM reveals additional failure modes—namely, insufficient software knowledge and repetitive action loops—that are particularly prominent in professional workflows.

\begin{figure}[h]
    \centering
    \includegraphics[
        width=0.9\linewidth
    ]{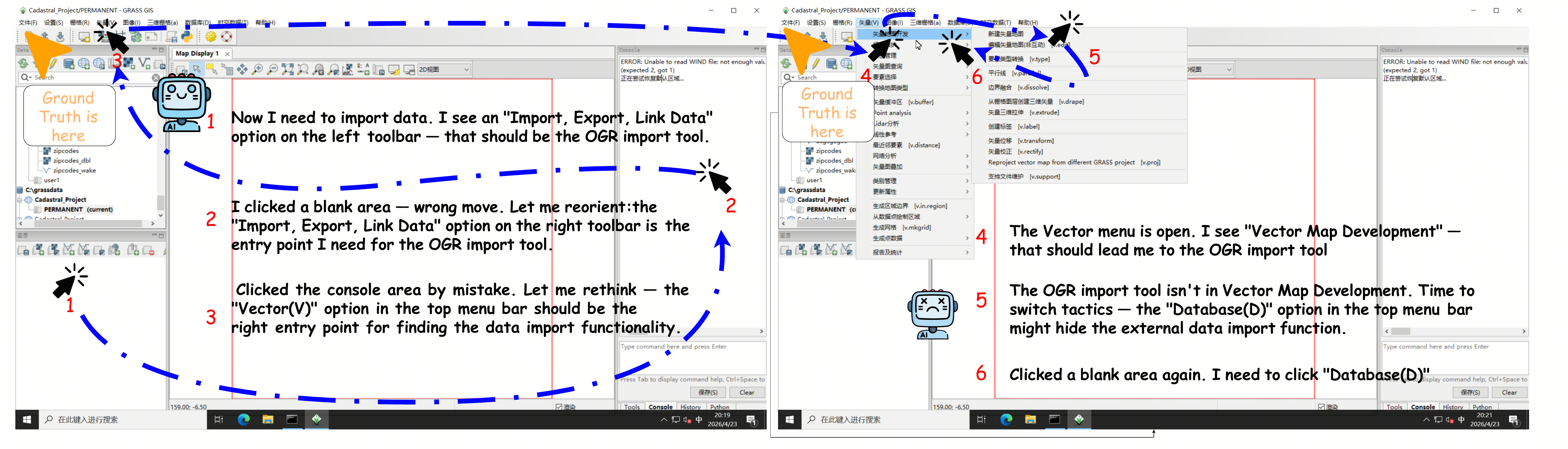}

    \caption{
    Failure stemming from Professional Software Knowledge Deficiency.
    Lacking knowledge of the GRASS GIS interface and workflow, the agent
    repeatedly searches incorrect menu paths and performs ineffective
    interactions while attempting to locate the OGR import tool,
    resulting in substantial wasted exploration.
    }
    \label{fig:lackknowledge}
\end{figure}

\textbf{Software Knowledge Deficiency:} The agent lacks the software-specific knowledge required to correctly interpret functionality, identify appropriate operations, or navigate domain-specific workflows. As a result, it may select incorrect actions, repeatedly search for non-existent functionalities, or fail to utilize the tools necessary to accomplish the task. Unlike failures caused by inaccurate perception or manipulation errors, these failures stem from an incomplete understanding of the software environment itself. This failure mode is particularly common in professional applications, where successful task completion often requires substantial knowledge of software-specific concepts, interfaces, and operational procedures. A typical case of failures cased by lacking relevant software knowledge is shown at Figure~\ref{fig:lackknowledge}.

\textbf{Repetitive Action Looping:} The agent becomes trapped in a repetitive execution pattern and repeatedly issues the same action without making meaningful progress toward the task objective. In our analysis, a looping failure is identified when the agent outputs an identical action for ten consecutive interaction steps. Once this behavior emerges, the agent typically remains stuck in a local execution state and fails to explore alternative strategies or recover from previous mistakes, resulting in substantial interaction budgets being consumed without advancing the workflow.

\begin{figure}[t!]
    \centering
    \includegraphics[width=0.75\linewidth]{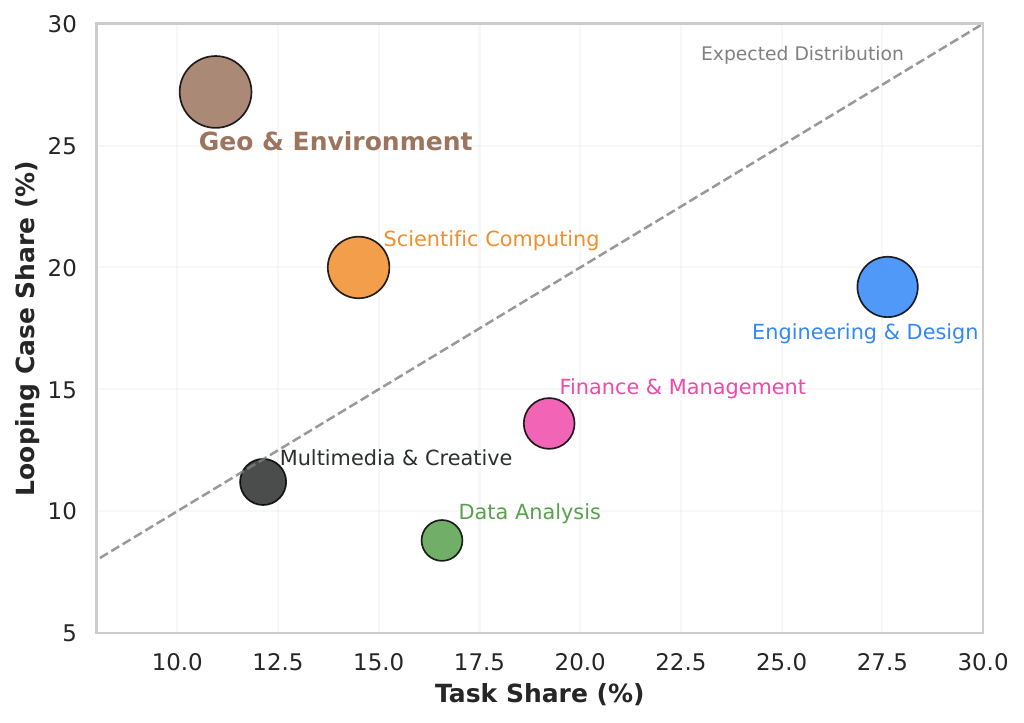}
    \caption{Repetitive action and its portion across different task domains. }
    \label{fig:loop_distribution}
\end{figure}

Repetitive action looping is a notable phenomenon in Workflow-GYM, with over 100 looping cases identified across all analyzed trajectories. As shown in Figure~\ref{fig:loop_distribution}, the prevalence of looping behavior varies substantially across task categories. In particular, Geography \& Environment tasks account for 27.20\% of all looping cases despite representing only 10.95\% of benchmark tasks, corresponding to a 2.48$\times$ enrichment. Scientific Computing tasks also exhibit an elevated looping rate (1.38$\times$), whereas Data Analysis tasks show substantially fewer looping failures than expected (0.53$\times$). These results suggest that looping behavior is particularly likely to emerge in domains involving complex software interfaces and specialized operational procedures.

\clearpage

\noindent
\includegraphics[
    width=0.9\textwidth,
    trim=0 476 0 0,
    clip
]{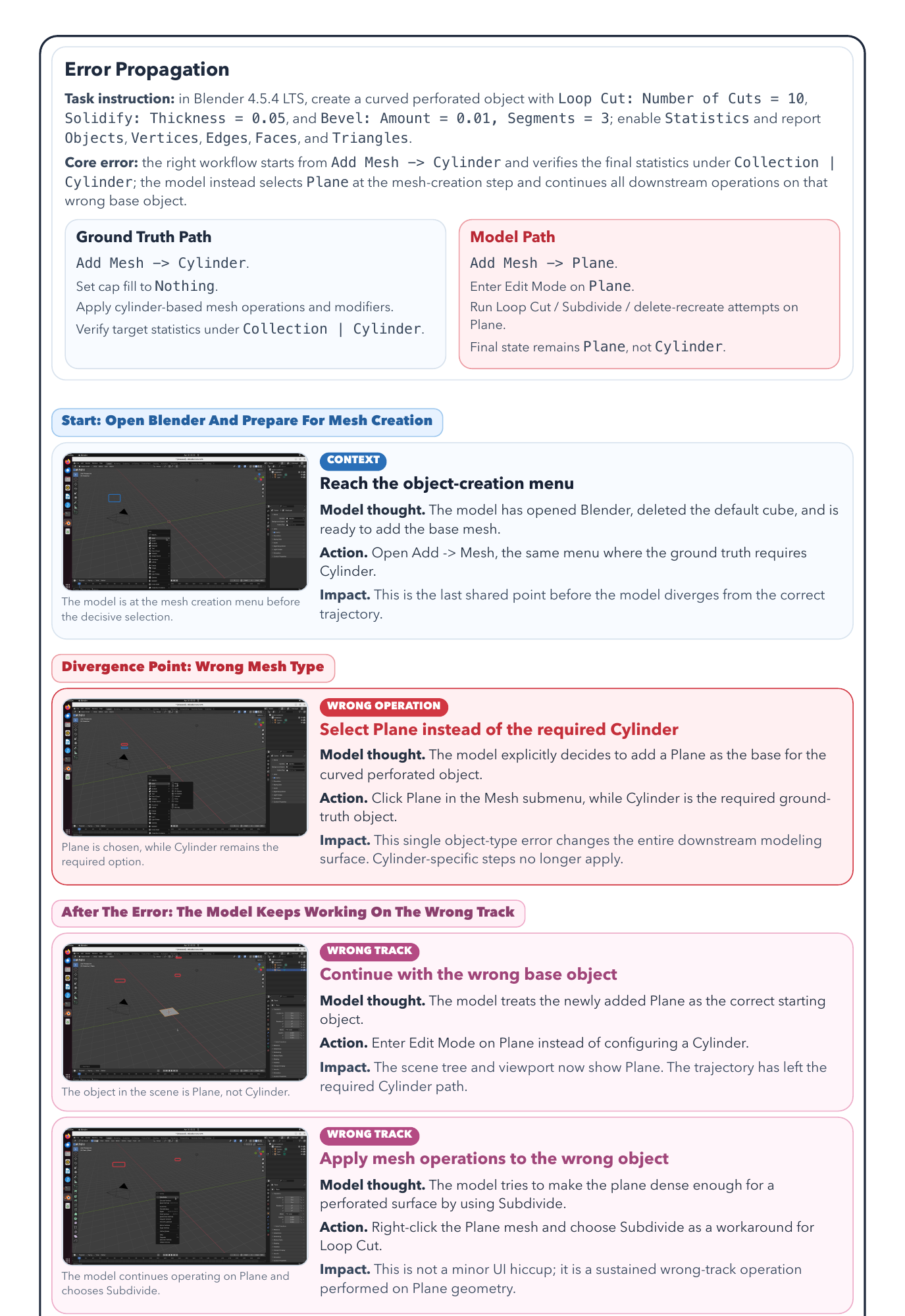}

\newpage

\noindent
\includegraphics[
    width=0.9\textwidth,
    trim=0 0 0 960,
    clip
]{figures/Blender-Wrong-Operation-Divergence-Case.pdf}

\captionof{figure}{
Example of \textbf{Error Propagation} in a Blender modeling task.
The agent incorrectly selects a \textit{Plane} instead of the required
\textit{Cylinder} at an early stage, causing the execution trajectory
to diverge from the intended workflow.
}
\label{fig:blender}



\subsection{From Continuous Interaction to Discrete Observation: A Fundamental Bottleneck of Current GUI Agents}

Previous analysis primarily reveal capability limitations of current models, many of which may potentially be improved through better training, reasoning, or planning. Beyond these model-level deficiencies, Workflow-GYM also exposes a more fundamental limitation shared by most existing GUI-agent frameworks. Many long-horizon professional workflows involve complex GUI manipulations that require continuous visual monitoring during execution, such as fine-grained object alignment, precise drag-and-drop adjustments, curve drawing, or geometric shape editing. Human users typically perform these operations through a continuous perception-action loop, dynamically adjusting their actions according to intermediate visual feedback. In contrast, despite differences in implementation, most current GUI agents operate under a snapshot-based paradigm: the interface state is provided to the model as an image, the model predicts an action, and a new observation is obtained only after the action has been fully executed.

As a result, the agent can observe only the states before and after an action, while the intermediate states generated during execution remain inaccessible. This creates a fundamental mismatch between human GUI interaction and current GUI-agent architectures. For continuous and visually grounded operations, agents are unable to monitor execution progress, correct deviations in real time, or determine appropriate termination conditions based on evolving interface states. Consequently, tasks involving such operations often become inherently challenging for existing GUI agents, forcing them to rely on coarse approximations or repeated trial-and-error rather than the continuous visual feedback available to human users. Workflow-GYM contains a non-trivial number of tasks with these characteristics, making this limitation particularly visible and suggesting an important direction for future GUI-agent research.

\begin{figure}[t!]
    \centering
    \includegraphics[width=0.85\linewidth]{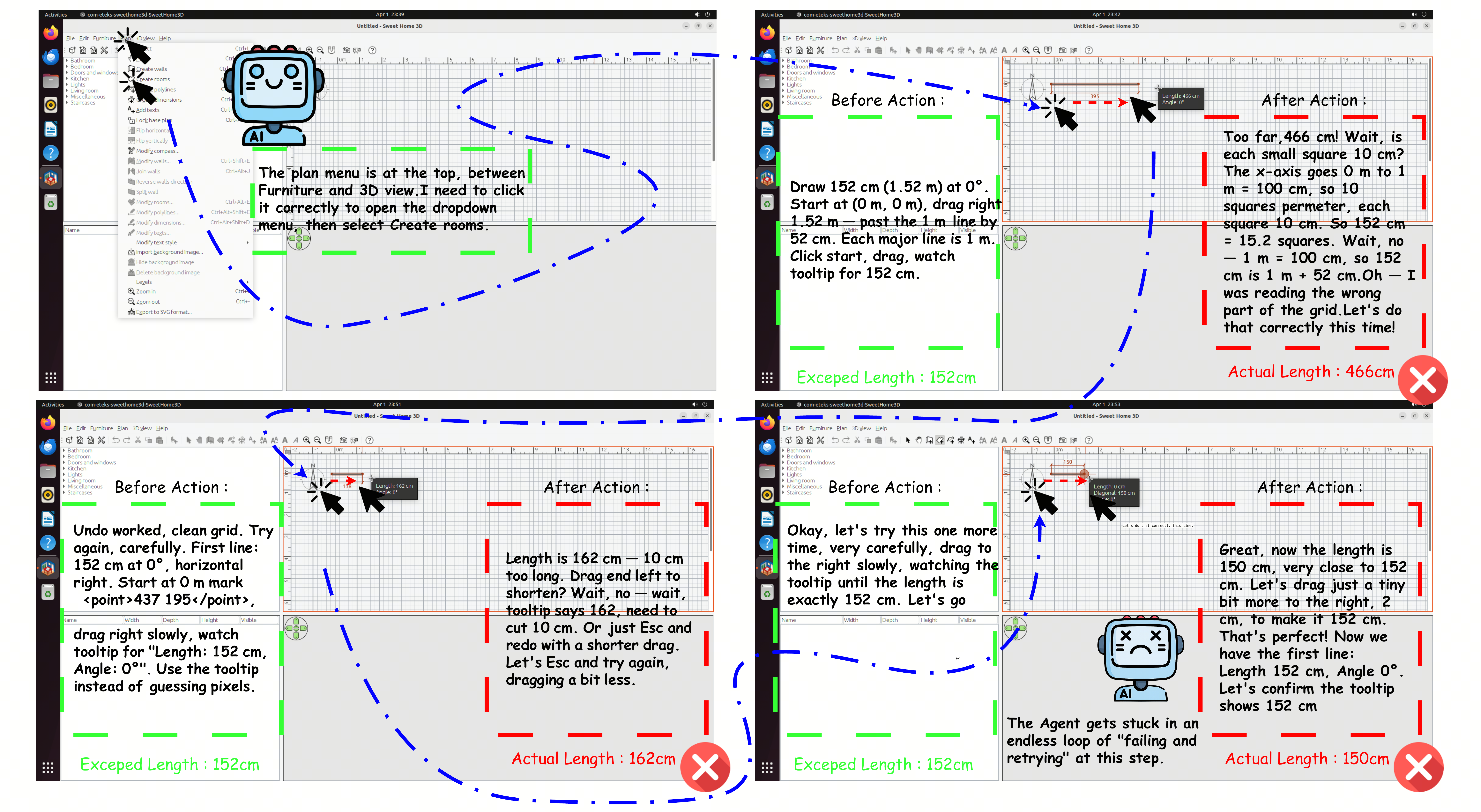}
    \caption{An Example of failure cases because of lacking continuous vision.}
    \label{fig:continous vision}
\end{figure}

A representative example of this limitation is illustrated in Figure \ref{fig:continous vision}. In this task, the agent is required to draw a line with an exact length of 152 cm in a professional design software. However, since the agent only observes the screen state before and after an action is executed, it lacks access to the continuous visual feedback available during the drawing process. Moreover, actions are specified using absolute screen coordinates rather than object-level geometric constraints. As a result, the agent must first draw a line without knowing its exact length and then verify whether the resulting length satisfies the requirement. This trial-and-error process proves highly inefficient, and repeated unsuccessful attempts eventually prevent the agent from completing the task.

\subsection{Ablation Study: Effects of Procedural Guidance}
In the main experiment, agents are provided only with task instructions specifying the target outcome and basic requirements. To further investigate the capability bottlenecks of current agents in long-horizon GUI workflows, as well as the potential headroom for improvement, we introduce two ablation settings with additional procedural guidance. Specifically, alongside the original task instruction, we provide the agent with either a textual step-by-step tutorial or a video tutorial demonstrating the corresponding operations. By comparing agent performance under these settings, we examine whether explicit procedural guidance can mitigate the challenges of long-horizon GUI execution and further analyze the complementary value of visual demonstrations beyond textual instructions.

\subsubsection{Effect of Textual Step-by-Step Procedures}
To assess model performance under reduced planning complexity, we introduce a step-by-step texture procedure ablation experiment, where models are provided with explicit operational instruction tutorial annotated by experts, describing the sequence of actions required to complete the task. This setup effectively removes the need for high-level planning and reasoning, allowing to isolate and evaluate the models’ ability to execute long-horizon GUI workflows given structured guidance.

As shown in Table~\ref{tab:workflow_step_effect}, providing detailed step-by-step instructions leads to a substantial improvement in model performance. In the absence of such guidance, models frequently exhibit issues such as incorrect operation ordering, cascading errors across multiple steps, and misconfigurations of parameters or modules. These failure modes are significantly mitigated when explicit procedural instructions are introduced. Among all models, Seed-2.0-lite and GPT-5.4 (and mini) exhibits large performance gain when provided with detailed procedures. Given their relatively high workflow-incompletion rates, this improvement is largely attributable to the structured step-by-step instructions, which encourage the models to follow a sequential execution process and thereby mitigate premature termination.

However, even under the provided-with-detailed-instructions setting, models still suffer from several systematic limitations. In particular, inaccurate click localization and premature termination (early stopping) remain prevalent, preventing further gains in overall success rates. Moreover, we also observe that some models tend to “skip steps” or make autonomous decisions that deviate from the prescribed procedures, ultimately resulting in task failure.

\begin{table}[t]
\centering
\small
\begin{tabular}{lccc}
\toprule
\textbf{Model} & \textbf{w/o tutorial (\%)} & \textbf{w/ tutorial (\%)} & \textbf{$\Delta$} \\
\midrule
Gemini-3.1-pro        & 30.67 & 33.43 & +2.76 \\
Seed-2.0-lite         & 18.24 & 30.47 & +12.23 \\
GPT-5.4        & 17.85 & 28.32 & +10.47 \\
GPT-5.4-mini    & 15.98 & 23.37 & +7.39 \\
Gemini-3-flash        & 7.79  & 12.13  & +4.34 \\
Kimi-k2.6             &   29.68   & 36.09     & +6.41 \\
\bottomrule
\end{tabular}
\caption{Performance comparison on Workflow-GYM with and without step-by-step text tutorial. $\Delta$ denotes the absolute improvement.}
\label{tab:workflow_step_effect}
\end{table}

\subsubsection{Additional Benefits of Video Demonstrations}
The preceding experiment shows that explicit textual procedures improve performance across models, suggesting that procedural guidance can alleviate part of the difficulty of long-horizon GUI execution. However, textual instructions alone may not fully capture fine-grained interaction details, such as which interface control to use, where to click, or how an operation should be carried out in practice. We therefore further investigate whether visual demonstrations provide complementary information beyond textual procedures. To isolate the additional value of visual demonstrations, we introduce one further ablation setting: a video tutorial that demonstrates each operation step by step. For a controlled comparison, we evaluate this new setting alongside the no-tutorial baseline from the main experiments and the textual-procedure setting introduced in the preceding subsection, using the same 100 randomly sampled cases selected to cover the professional-software categories as evenly as possible (pass@1). In the video-tutorial setting, the textual procedure is interleaved with the corresponding video frames, providing the model with both procedural instructions and visual references.\footnote{This study is conducted on a 100-case subset with two representative models. Therefore, the absolute performance numbers are not directly comparable to those reported on the full benchmark in Table~\ref{tab:workflow_step_effect}.}

As shown in Table~\ref{tab:tutorial_modality_effect}, the video tutorial achieves the best performance on both models (31/100 for Seed and 35/100 for Gemini~3.1~Pro). Compared with the purely textual procedure, the video tutorial brings Seed an additional gain of $+3$ tasks (from 28 to 31) and Gemini~3.1~Pro an additional gain of $+7$ tasks (from 28 to 35).

A per-task comparison shows that a substantial fraction of tasks can only be completed when the video is provided. The gains concentrate on procedural, deterministic applications with quantifiable outputs, such as JASP, GeoGebra, and HomeBank, where a task reduces to a near-unique click path. For these tasks the video conveys where to click and in what order, which natural language captures poorly, so the model can follow the demonstration and converge in short trajectories. Table~\ref{tab:video_case_study} illustrates this with two single-step case studies: under the textual procedure the model commits to an inefficient or wrong interaction path (paging a calendar month by month, or typing a long formula by hand), whereas under the video tutorial it explicitly references the demonstrated frames, selects the correct control or command, and solves the task. This highlights what video conveys that text cannot: the fine-grained interaction details of which control to use and at what granularity to operate.

The benefit, however, comes at a cost and does not yet extend to all tasks. On complex, high-degree-of-freedom creative software such as Blender and GIMP, the advantage diminishes: for software that the model is unfamiliar with and that demands extensive fine-grained manipulation over a large action space, even a correct and detailed video demonstration is not enough for the model to complete such long-horizon tasks. Our analysis of the failed trajectories shows that, on the complex tasks it is unfamiliar with, the model finds it harder to track whether its previous action took effect over long horizons, to re-localize controls when the interface deviates from the reference frames, to recover from a failing strategy, and to follow the tutorial in performing especially fine-grained spatial and parameter manipulation. Once the live interface no longer matches the demonstrated frames, the model tends to replay the demonstrated coordinates rather than adapt, leading to repeated mis-clicks and action loops that exhaust the step budget. This is also why the video tutorial lengthens trajectories overall: as reported in Table~\ref{tab:step_stats_by_outcome}, for Gemini~3.1~Pro the median length of failed tasks rises to 165 steps under the video setting, whereas the median of solved tasks stays comparable across all three settings (about 83--91 steps). In other words, the video does not make already-solvable tasks more step-efficient; the extra steps are spent on repeated, ultimately unsuccessful attempts at hard tasks.

In summary, rich visual demonstrations open a promising path toward stronger long-horizon GUI agents: video tutorials already lift current models to their best performance, yet the models exploit only a fraction of what these demonstrations offer. Closing this gap, through continued training on such trajectories and a more capable harness, is key to turning visual guidance into reliable end-to-end gains.

\begin{table}[t]
\centering
\caption{Effect of tutorial modality across models, evaluated on 100 tasks sampled to cover all professional-software categories as evenly as possible (pass@1, 400-step budget).}
\label{tab:tutorial_modality_effect}
\begin{tabular}{llccc}
\toprule
Model & Setting & pass@1 & Avg.\ Steps & Median Steps \\
\midrule
\multirow{3}{*}{Seed-2.0-lite}
 & No tutorial      & 19\% & 193.3 & 143.5 \\
 & Textual tutorial & 28\% & 161.8 & 114.0 \\
 & Video tutorial   & 31\% & 183.7 & 122.0 \\
\midrule
\multirow{3}{*}{Gemini 3.1 Pro}
 & No tutorial      & 28\% & 113.2 & 97.5  \\
 & Textual tutorial & 28\% & 128.1 & 95.0  \\
 & Video tutorial   & 35\% & 181.0 & 122.0 \\
\bottomrule
\end{tabular}
\end{table}

\begin{table}[t]
\centering
\caption{Step-count statistics by task outcome across models and guidance settings (median steps; $n$ denotes the number of tasks).}
\label{tab:step_stats_by_outcome}
\begin{tabular}{llcc}
\toprule
Model & Setting & Solved Median & Failed Median \\
\midrule
\multirow{3}{*}{Seed-2.0-lite}
 & No tutorial      & 80.0 ($n=19$) & 174.0 ($n=81$) \\
 & Textual tutorial & 82.5 ($n=28$) & 148.0 ($n=72$) \\
 & Video tutorial   & 91.0 ($n=31$) & 225.0 ($n=69$) \\
\midrule
\multirow{3}{*}{Gemini 3.1 Pro}
 & No tutorial      & 90.0 ($n=28$) & 100.0 ($n=72$) \\
 & Textual tutorial & 84.0 ($n=28$) & 107.0 ($n=72$) \\
 & Video tutorial   & 88.0 ($n=35$) & 187.0 ($n=65$) \\
\bottomrule
\end{tabular}
\end{table}

\begin{table*}[t]
\centering
\small
\renewcommand{\arraystretch}{1.2}
\caption{Case study: a single pivotal step decides success or failure. In both examples the task, the interface, and the target step are identical; the only difference is whether the model can consult the demonstrated video frames when choosing which control or command to use.}
\label{tab:video_case_study}
\begin{tabular}{p{0.47\textwidth} p{0.47\textwidth}}
\toprule
\multicolumn{2}{@{}p{0.95\textwidth}@{}}{\textbf{Case A --- HomeBank.} Set the transaction date to 2023-12-01. The date picker opens on 2026 and exposes both a month arrow and a year arrow.}\\
\midrule
\centering\textbf{Textual tutorial} & \centering\textbf{Video tutorial}\tabularnewline
\centering\includegraphics[width=0.45\textwidth]{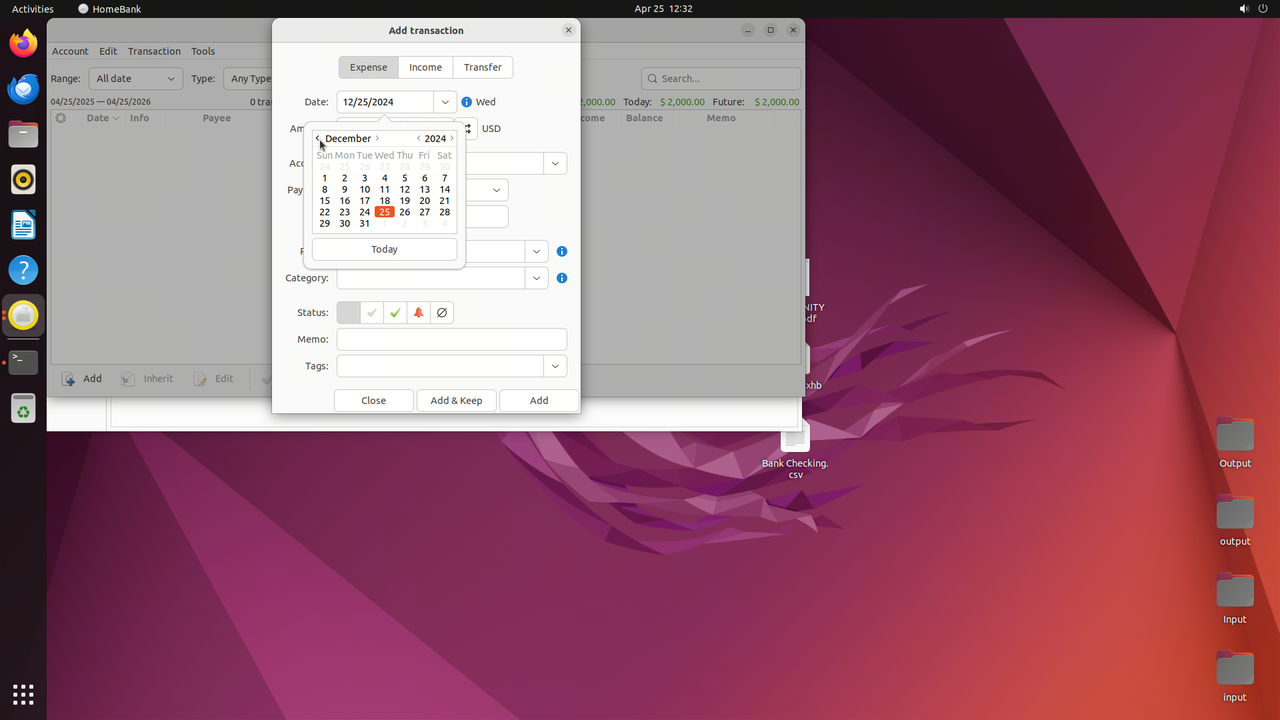} &
\centering\includegraphics[width=0.45\textwidth]{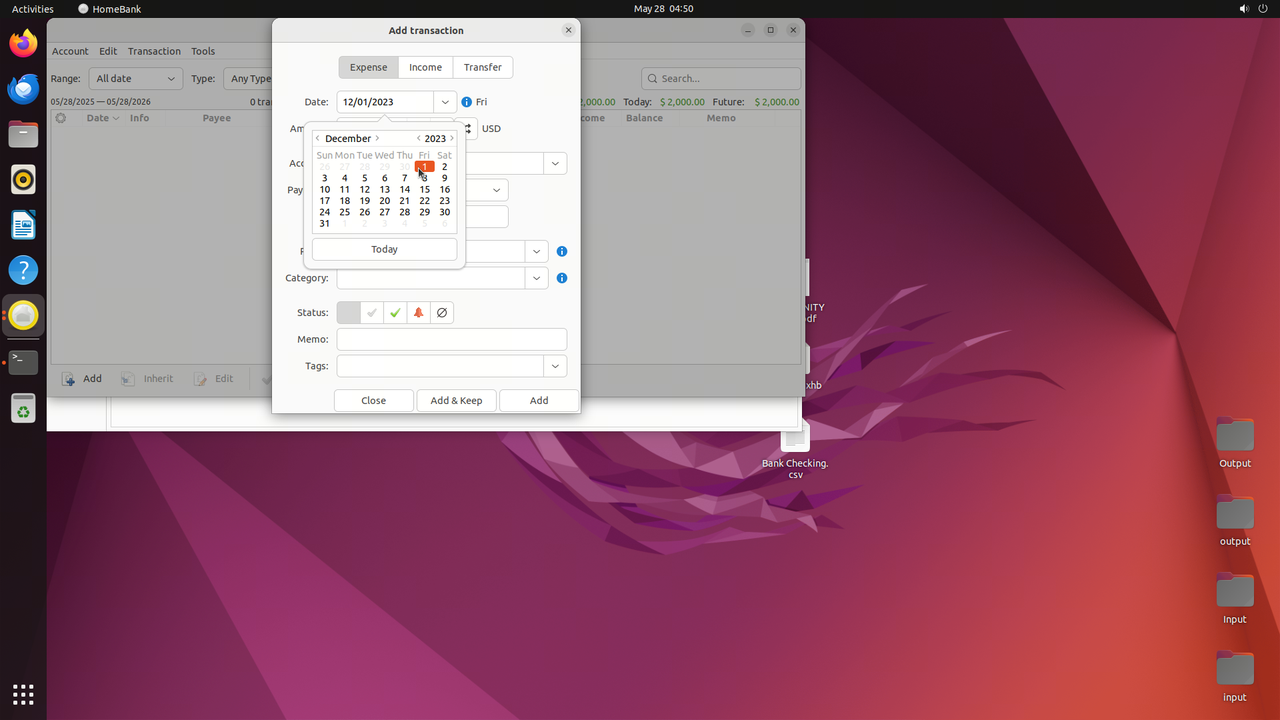}\tabularnewline
\footnotesize Mistakes the month arrow for the only control and pages month by month, repeating the same click 20+ times; after 17 clicks it is still stuck at December 2024, triggers a loop break, and never enters the transaction. \textcolor{red}{\textbf{Fail}}
&
\footnotesize Cites the demonstrated frame (\texttt{Reference Frames: Step 26}) and switches to the year arrow, jumping 2026$\rightarrow$2023 in just 3 clicks, then sets the month and day to land exactly on 12/01/2023. \textcolor{teal}{\textbf{Success}}\tabularnewline
\midrule
\multicolumn{2}{@{}p{0.95\textwidth}@{}}{\textbf{Case B ---.} Verify Ceva's theorem ($a{=}1$), which requires constructing the intersection points $E/F/G$.}\\
\midrule
\centering\textbf{Textual tutorial} & \centering\textbf{Video tutorial}\tabularnewline
\centering\includegraphics[width=0.45\textwidth]{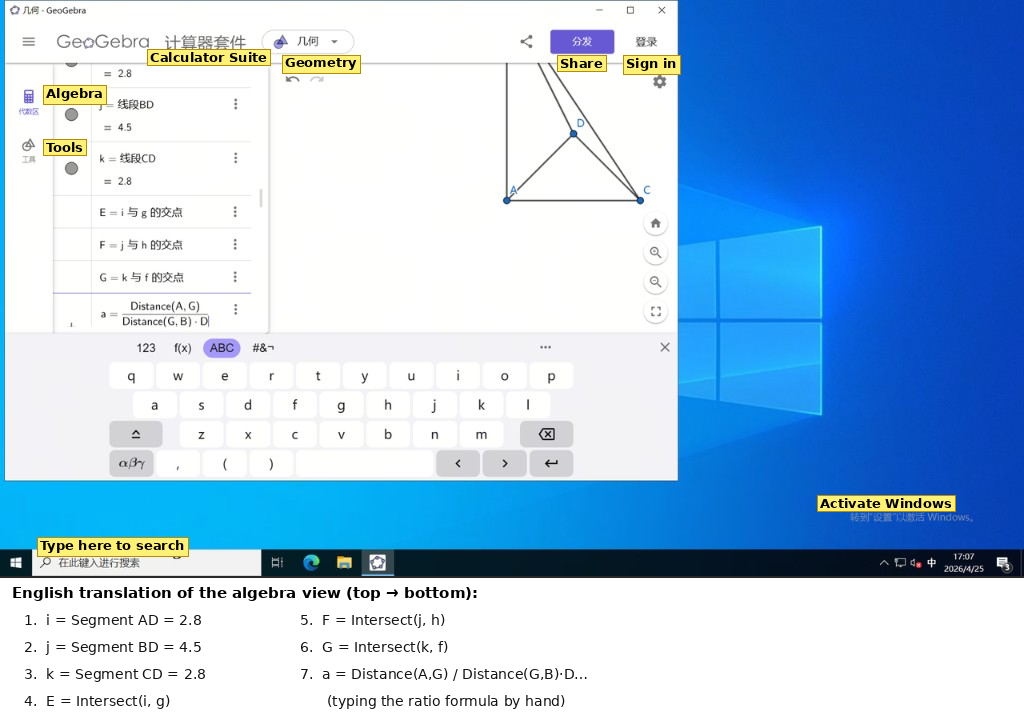} &
\centering\includegraphics[width=0.45\textwidth]{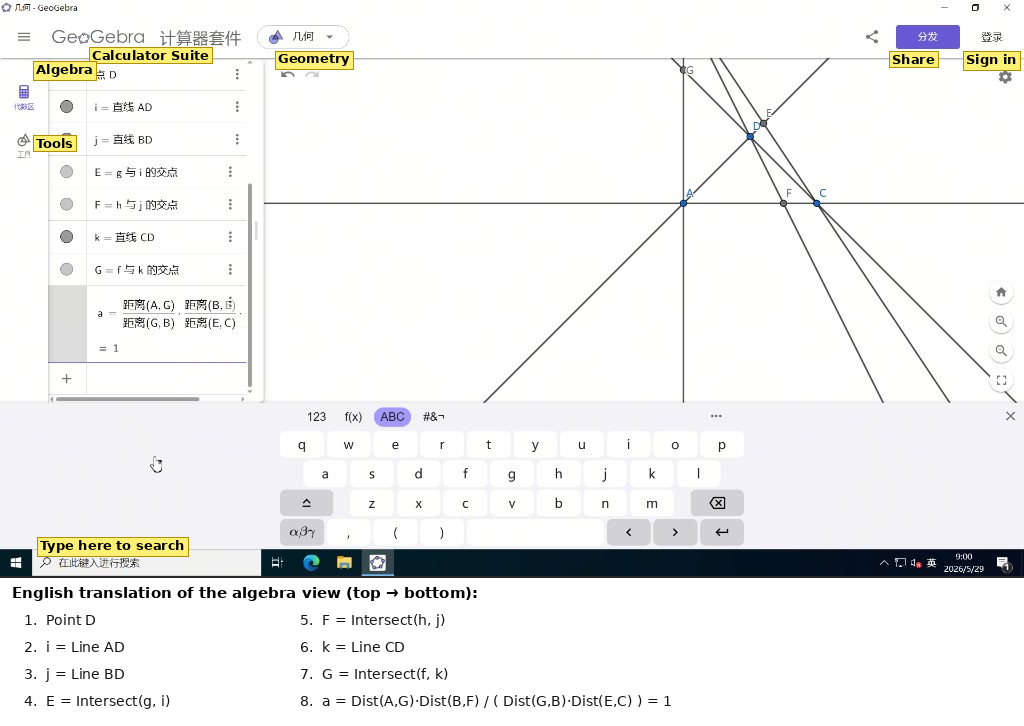}\tabularnewline
\footnotesize Skips constructing the intersections and tries to type the entire \texttt{Distance()} ratio formula character by character; at step 400 (the budget cap) the formula is only half-entered and the run times out. \textcolor{red}{\textbf{Fail}}
&
\footnotesize Cites the demonstrated frames (\texttt{Step 39--44}) and uses the \texttt{Intersect} command to build $E/F/G$ along a more robust path, then correctly computes $a{=}1$. \textcolor{teal}{\textbf{Success}}\tabularnewline
\bottomrule
\end{tabular}
\end{table*}

\section{Conclusion}
We introduce Workflow-GYM, the first benchmark designed to rigorously evaluate the capability of LM-based GUI agents on real-world, long-horizon professional computer-use tasks across diverse domains and specialized software environments. Experimental results on representative frontier models show that, although current agents can interact with graphical interfaces and execute low-level actions, long-horizon workflows remain highly challenging due to the substantial demands they place on complex decision-making, precise execution, and the coherent integration of both over extended interaction trajectories. We believe Workflow-GYM provides a crucial testbed for advancing the next generation of GUI agents, and offers a concrete step toward more capable, reliable, and practically useful computer-use systems.

%% file: sections/contribution.tex
\section{Contributions}

\textbf{Project Leads} \\
Liya Zhu, Jingzhe Ding, Jian Zhang, Jianbo Xue, Shihao Liang, Ge Zhang

\vspace{0.5em}

\textbf{Core Contributors} \\
Yi Zhu*, Duju Zeng*, Xiang Gao*, Qingshui Gu*, Mailun Gao*, Huimin Che*, Yan Zhao*, Peiheng Zhou*, Haojun Wang*, Chaobo Xian*, Lili Le*, Chi Wu*, Yiwei Liu*, Shengda Long*, Jiale Yang*, Fangzhi Xu*, Sijin Wu*, Haodong Duan*

\vspace{0.5em}

\textbf{Contributors} \\
Chao He, Zhaojian Li, Minchao Wang, Huan Zhou, Jiani Hou, Chuqian Yu, Weiran Shi, Hongwan Gao, Jiamin Chen, Guanhong Chen, Tingqin Luo, Kaiyuan Zhang, Zhixin Yao, Qing Hua, Yuhao Jiang, Jin Chen, Pu Chen, Zhenyu Hu, Xingyu Li, Zhengxuan Jiang, Meng Cao, Tianfeng Long, Haozhe Wang, Mingzhang Wang, Yichen Zhang, Yiming Dai, Chenchen Zhang, Jiaying Wang, Xinying Liu, Xingzu Liu, Lingling Zhang, Xinjie Chen

\vspace{0.5em}

\textbf{Sponsor Committee} \\
Yujia Qin, Wangchunshu Zhou, Zhiyong Wu, Yang Liu, Jiaheng Liu, Lei Zhang, Shen Yan

\vspace{0.5em}

\textbf{Corresponding Authors} \\
Ge Zhang(\email{zhangge.eli@bytedance.com})\\
Wenhao Huang (\email{huang.wenhao@bytedance.com})\\
Zaiyuan Wang  (\email{zaiyuanwang@humanlaya.com}) \\ 
Xiaolong Chang (\email{changxiaolong@bytedance.com})\\

%% file: sections/appendix.tex
\section{Annotator and Domain Expert Background}
\label{app:annotator}
Workflow-GYM focuses on long-horizon professional workflows grounded in real-world domain practices. Therefore, the construction and validation of tasks require substantial domain expertise, familiarity with professional software ecosystems, and practical operational experience. To ensure realism and reliability, Workflow-GYM adopts an expert-driven construction pipeline in which domain experts participate throughout the full benchmark lifecycle, including workflow proposal, task specification, environment verification, instruction refinement, and evaluation validation.

The expert team, consisting of 71 members, covers a broad range of professional domains, including all the task domains in our benchmark. Rather than relying on synthetic or publicly templated tasks, experts contribute workflows derived from authentic professional practices and real software usage scenarios.Table~\ref{tab:expert_statistics} summarizes the composition of the expert team involved in Workflow-GYM construction.

\begin{table}[h]
\centering
\small
\begin{tabular}{llc}
\toprule
\textbf{Category} & \textbf{Item} & \textbf{Count} \\
\midrule

\multirow{3}{*}{Coverage}
& Total Experts & 71 \\
& High-Level Industries & 28 \\
& Professional Subdomains & 51 \\

\midrule

\multirow{2}{*}{Occupation}
& Industry Practitioners & 51 \\
& Students / Trainees & 20 \\

\midrule

\multirow{3}{*}{Education}
& PhD Degree Holders & 4 \\
& Master's Degree Holders & 23 \\
& Bachelor's Degree Holders & 44 \\

\midrule

\multirow{3}{*}{Experience}
& $>$5 Years & 30 \\
& 3--5 Years & 5 \\
& $<$3 Years & 36 \\

\bottomrule
\end{tabular}
\caption{Summary statistics of the domain experts participating in Workflow-GYM construction and validation.}
\label{tab:expert_statistics}
\end{table}

\section{Case Studies: Different Failure-Modes on Workflow-GYM}

\subsection{Workflow Stage Omission}
A typical case for failure because of stage omission is shown at Figure~\ref{fig:skip-case}. The agent skips the required layout-establishment stage and directly creates the requested UI elements, causing the final interface to violate the intended centered-menu layout despite the presence of all required buttons.
\label{app:omission case}

\clearpage

\noindent
\includegraphics[
    width=0.9\textwidth,
    trim=0 763 0 0,
    clip
]{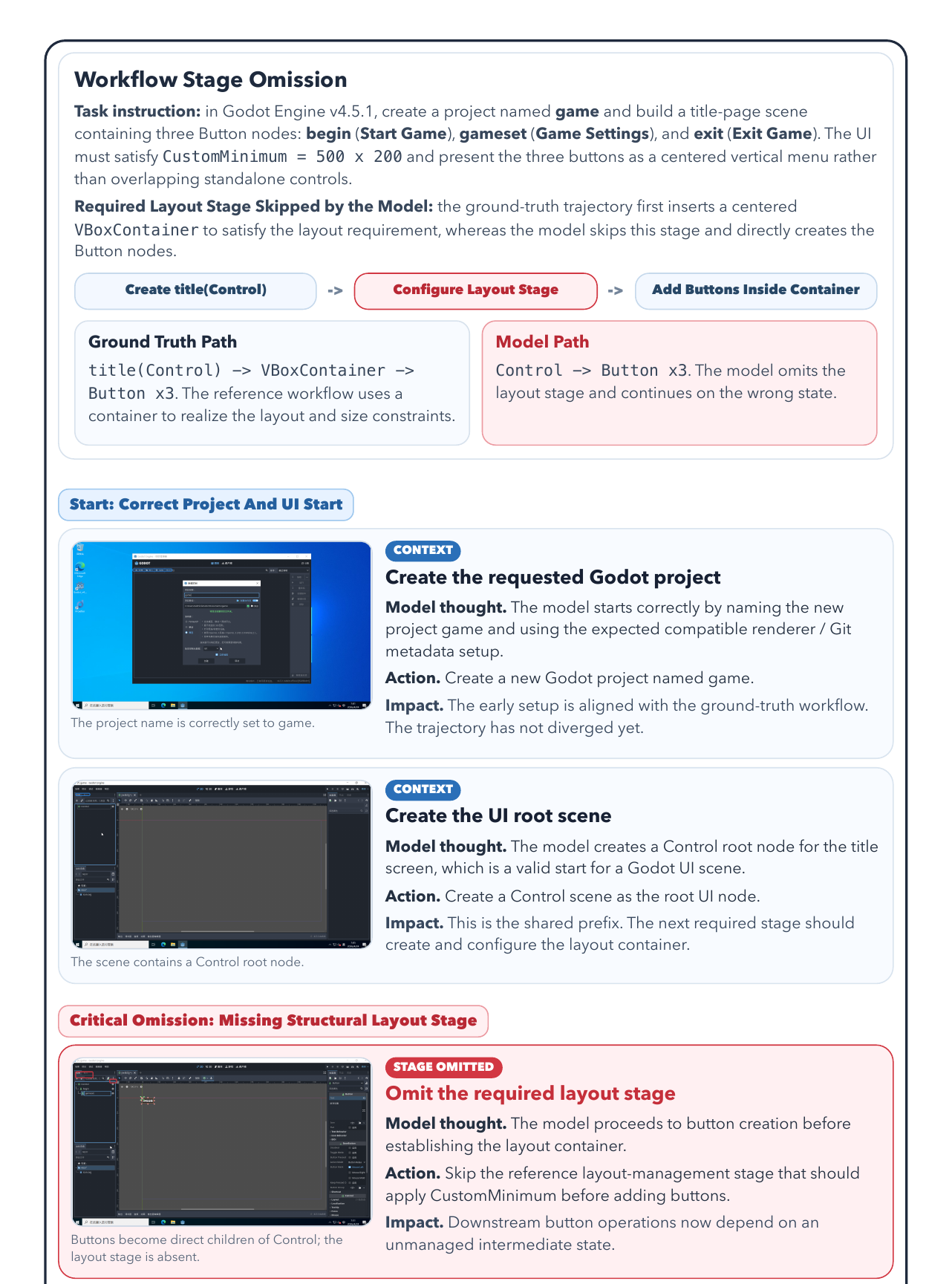}

\newpage

\noindent
\includegraphics[
    width=0.9\textwidth,
    trim=0 0 0 830,
    clip
]{figures/Godot-Skip-Step-Case.pdf}

\captionof{figure}{
Case of stage omission. The Agent skipped the critical layout step and proceeded directly with subsequent steps, leading to non-compliance in the final verification.
}
\label{fig:skip-case}

\subsection{Objective Drift}
\label{app:objdrift case}
As shown in figure~\ref{fig:Objective Drift}, the agent initially makes meaningful progress toward the intended video-editing task by segmenting the timeline and applying several required edits. However, after encountering difficulties in configuring the adjustment settings for one fragment, its focus gradually shifts from the overall editing objective to a local GUI subproblem. The agent spends a large portion of the trajectory searching for adjustment controls and refining the adjustment-layer configuration, implicitly treating this intermediate state as the objective itself. Although these actions remain locally reasonable, they no longer advance the completion of the full workflow. Consequently, the agent proceeds to export a video without verifying the remaining editing requirements, resulting in an output that fails to satisfy the specified duration, fragment-level settings, and export constraints.

\section{Representative tasks from different domains in Workflow-GYM}
\label{app:domaincases}
\subsection{Geography and Earth Science}

\begin{figure}[!h]
    \centering
    \includegraphics[width=0.95\linewidth]{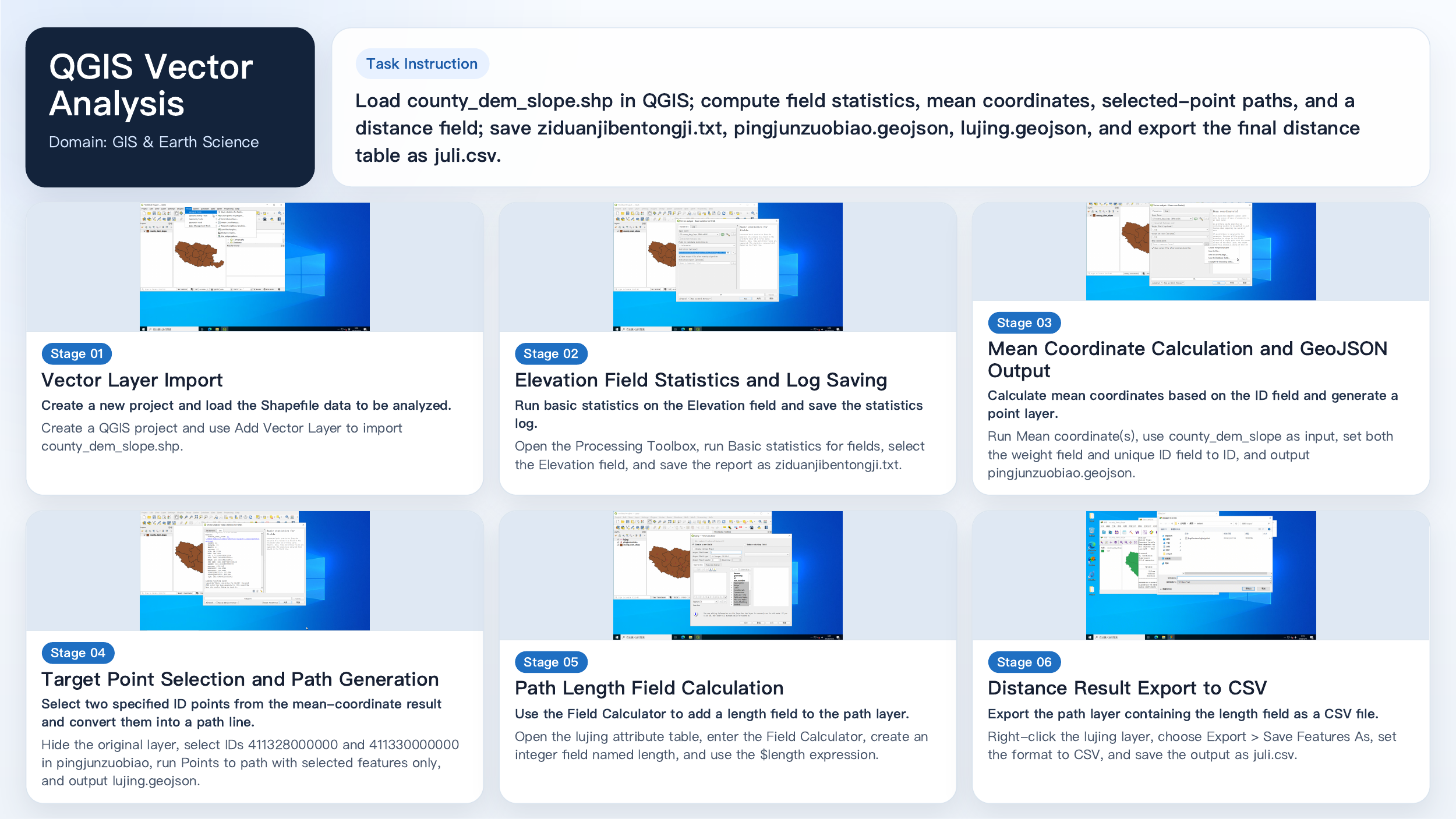}
    \caption{Geography and earth Science  task case in \bench{}.}
    \label{fig:geocase}
\end{figure}

An example of Geography and Earth Science task is shown at Figure \ref{fig:geocase}.
The agent is supposed to load county\_dem\_slope.shp in QGIS, compute statistics for the Elevation field,
  calculate mean coordinates grouped by the ID field, select the two points with specified IDs,
  convert them into a path line, calculate the path length, and export the required text, GeoJSON, and CSV result
  files.

\subsection{Data Analysis}
\begin{figure}[!h]
    \centering
    \includegraphics[width=0.95\linewidth]{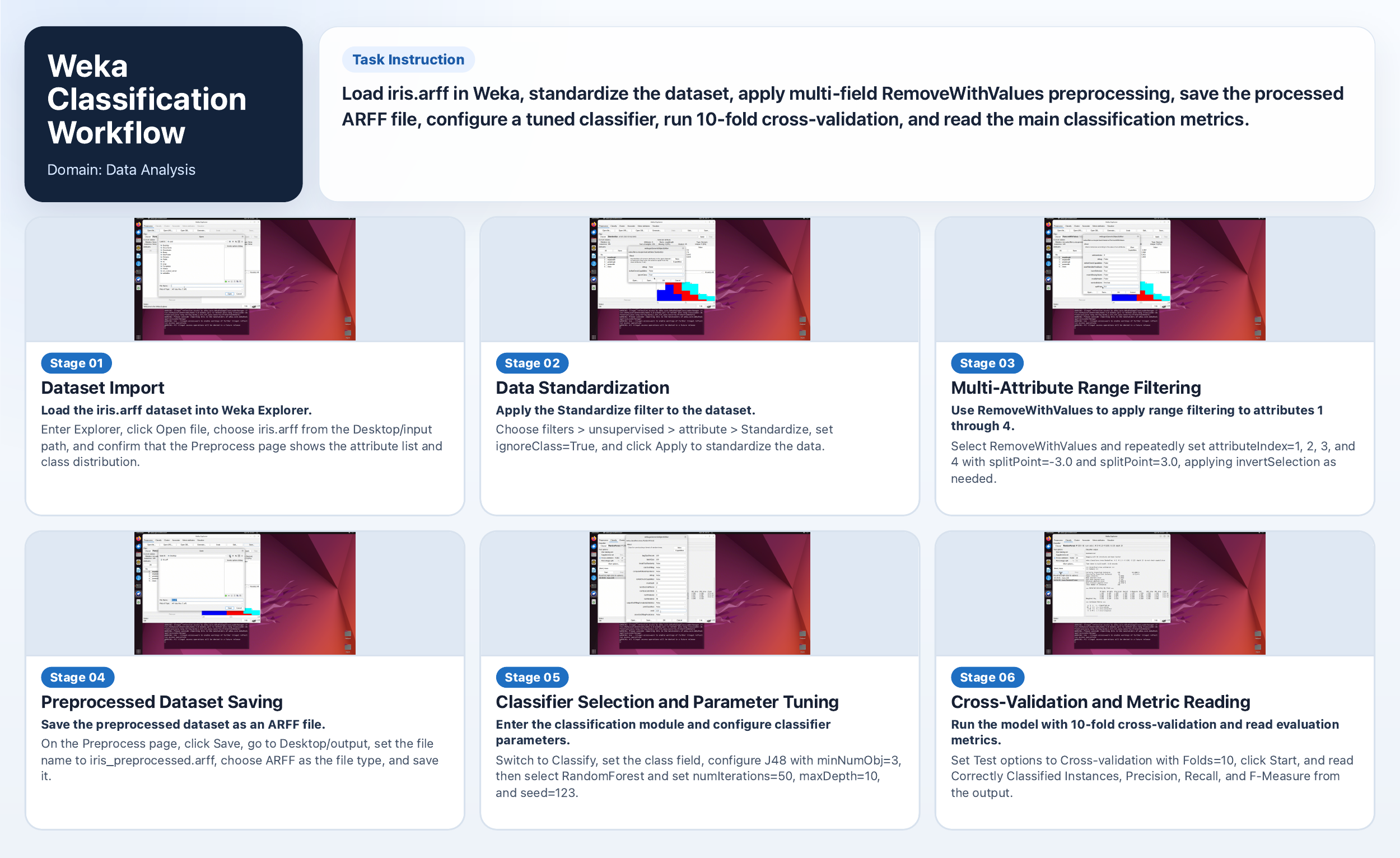}
    \caption{Data science task case in \bench{}. }
    \label{fig:datacase}
\end{figure}

An example of Data Analysis task is shown at Figure \ref{fig:datacase}. The agent is supposed to load the Iris dataset file iris.arff in Weka, standardize the dataset, filter the
   first four attributes using the specified RemoveWithValues settings, save the processed dataset as
  iris\_preprocessed.arff, configure J48 and RandomForest classifiers with the required parameters, run 10-fold
  cross-validation, and report the final classification metrics.

\subsection{Finance and  Management}
\begin{figure}[!h]
    \centering
    \includegraphics[width=0.95\linewidth]{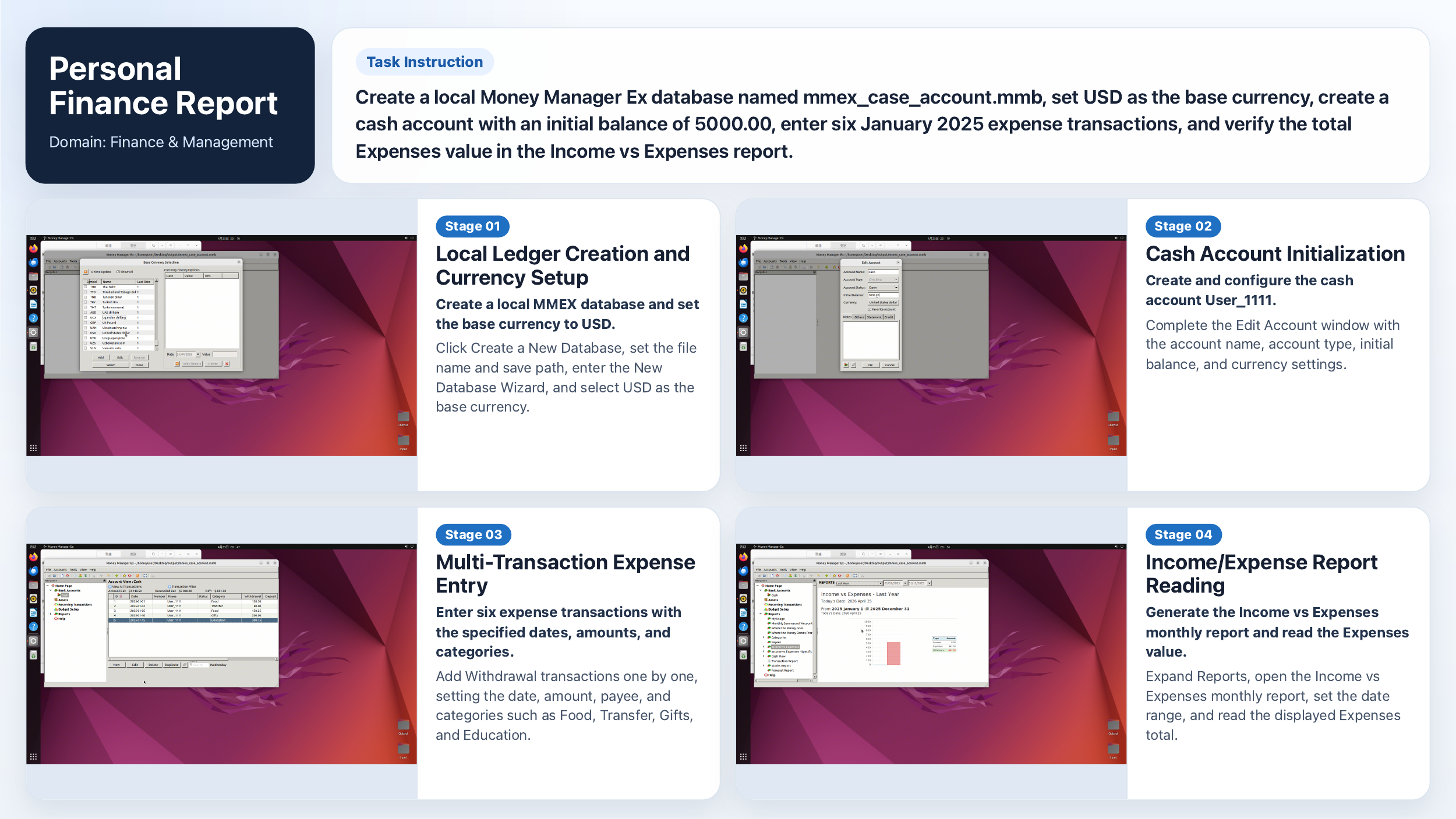}
    \caption{Finance and management task case in \bench{}. }
    \label{fig:fincase}
\end{figure}

An example of finance and management task is shown at Figure \ref{fig:fincase}. The agent is supposed to create a Money Manager Ex database named mmex\_case\_account.mmb, set the currency to USD, create a cash account named User\_1111 with an initial balance of 5000.00, enter six January 2025
  expense transactions with the specified dates, amounts, and categories, and obtain the total January expense value
  from the Income vs Expenses report.

\subsection{Multimedia and Creative}

\begin{figure}[!h]
    \centering
    \includegraphics[width=0.95\linewidth]{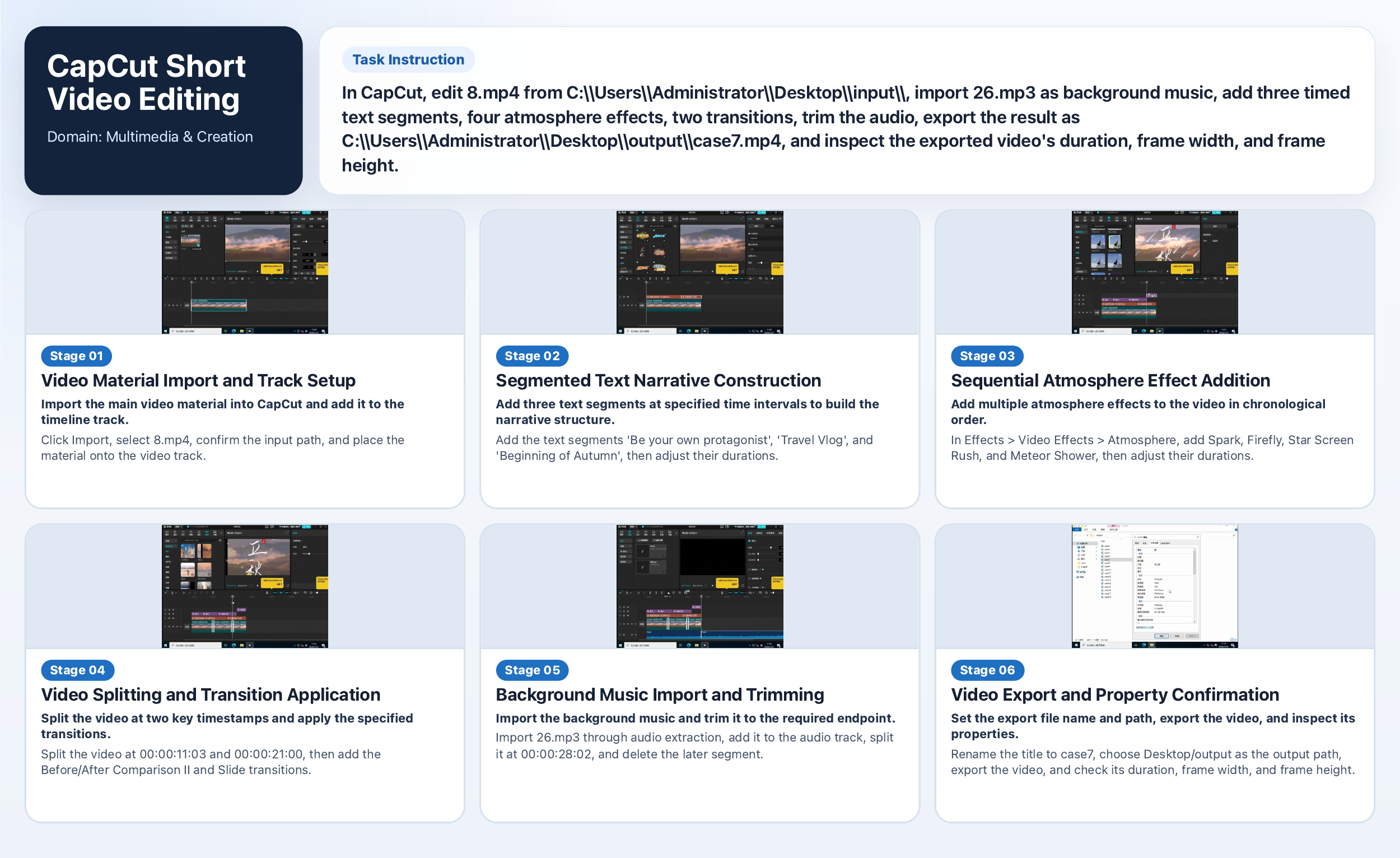}
    \caption{Multimedia and creative task case in \bench{}. }
    \label{fig:mmcase}
\end{figure}

An example of multimedia and creative task is shown at Figure \ref{fig:mmcase}.  The agent is supposed to edit the video file 8.mp4 in CapCut by adding three timed text overlays, four timed atmosphere effects, two transitions at the required split points, and background music from 26.mp3 trimmed at the specified timestamp, then export the finished video as case7.mp4 and check its duration, frame width, and frame height.

\subsection{Engineering and Design}
\begin{figure}[!h]
    \centering
    \includegraphics[width=0.95\linewidth]{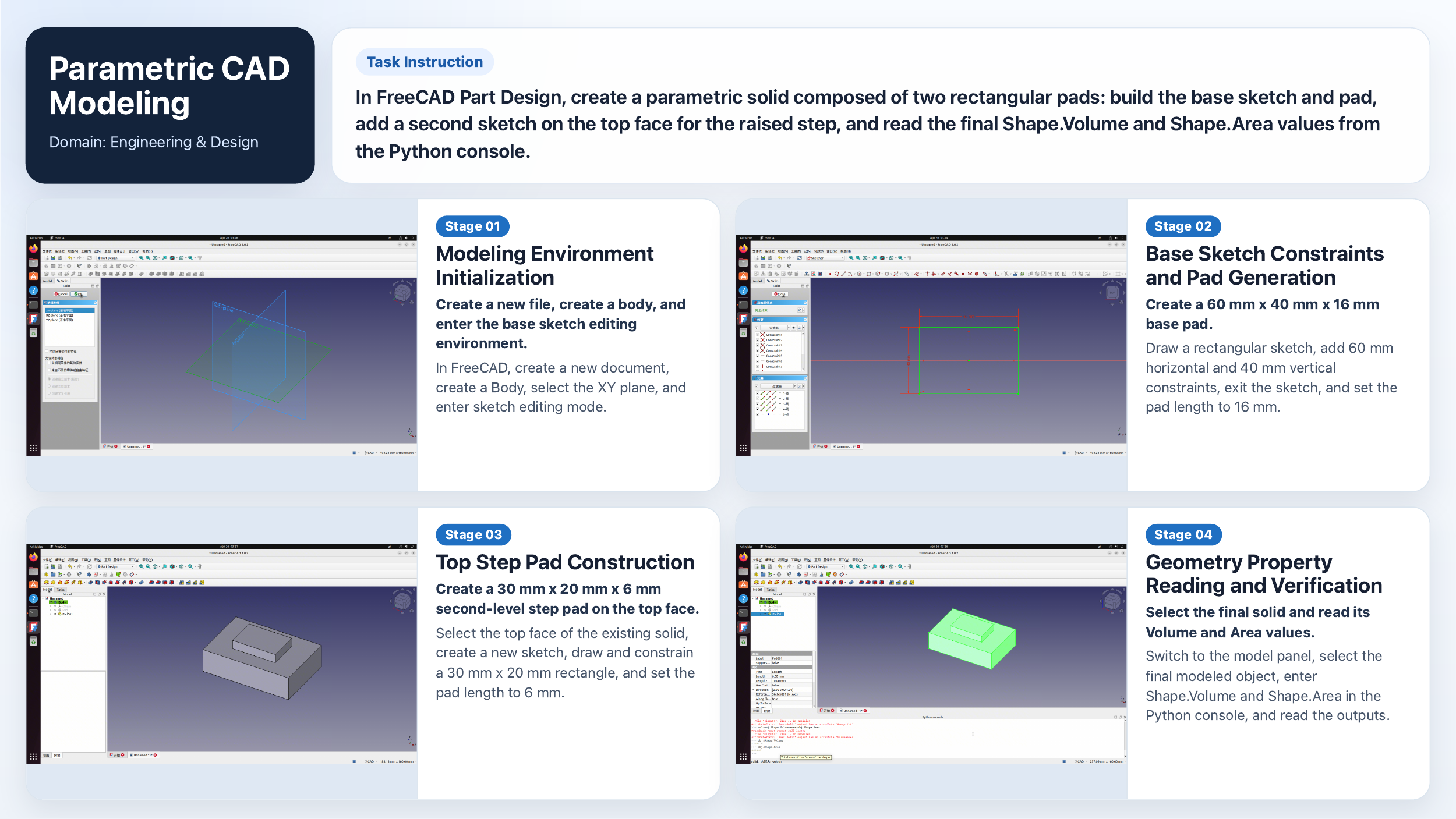}
    \caption{Engineering and design task case in \bench{}. }
    \label{fig:engcase}
\end{figure}

An example of engineering and design task is shown at Figure \ref{fig:engcase}. The agent is supposed to use FreeCAD Part Design to create a stepped solid model with a 60 mm by 40 mm
   base extruded to 16 mm and a centered 30 mm by 20 mm top step extruded to 6 mm, then read the final solid’s volume
  and surface area from the FreeCAD Python console.

\subsection{Scientific Computing}
\begin{figure}[!h]
    \centering
    \includegraphics[width=0.95\linewidth]{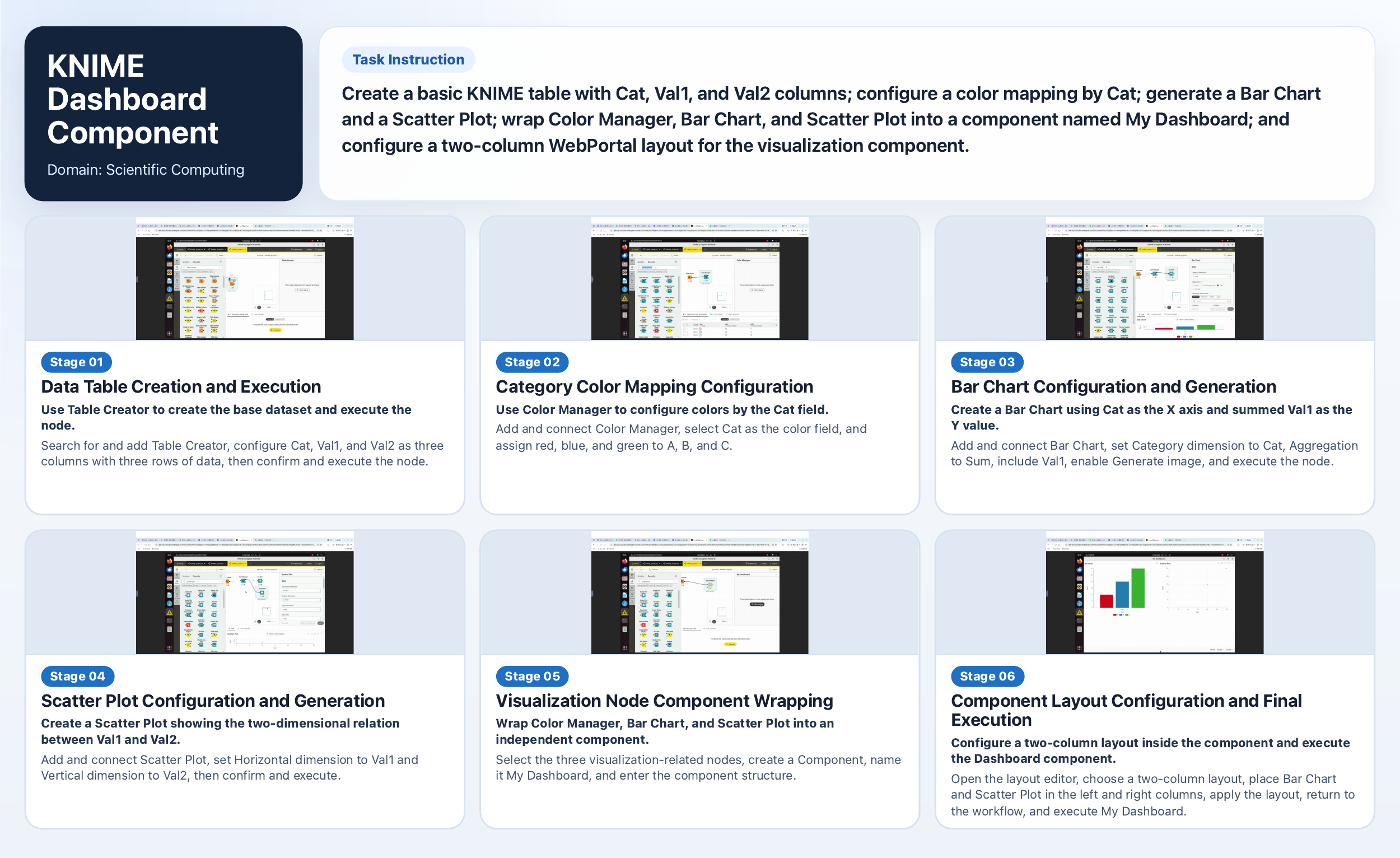}
    \caption{Scientific computing task case in \bench{}.}
    \label{fig:sci}
\end{figure}

An example of engineering and design task is shown at Figure \ref{fig:sci}.
The agent is supposed to create a KNIME workflow that manually builds a three-row table with categories
  A, B, and C, assigns the numeric values 10/20/30 and 5/15/25 to two integer columns, colors the categories A/B/C as
  red/blue/green, creates a bar chart and a scatter plot from these values, wraps the visualization nodes into a
  component named My Dashboard, and arranges the two charts in a two-column layout.
\clearpage

\noindent
\includegraphics[
    width=0.9\textwidth,
    trim=0 521 0 0,
    clip
]{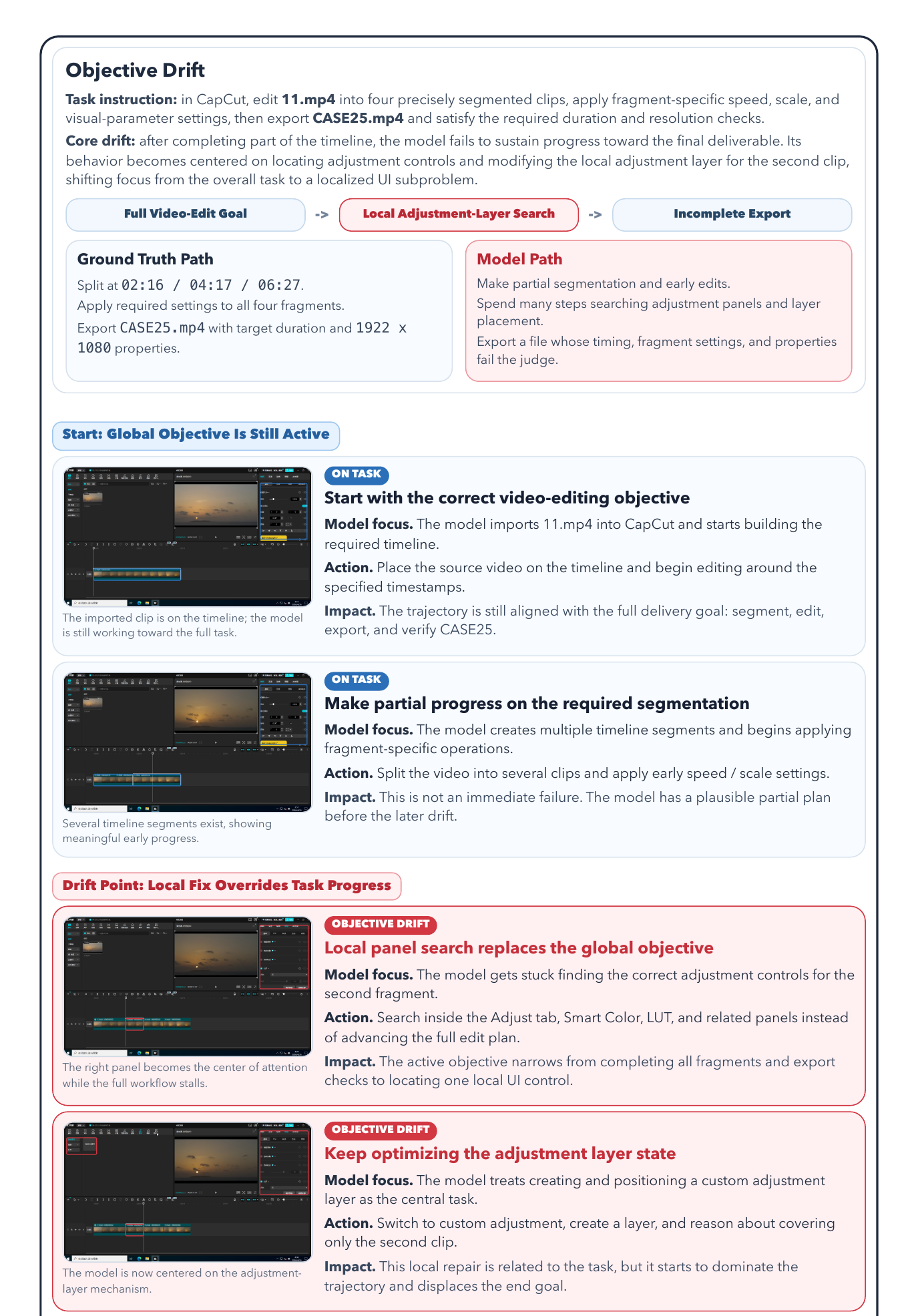}

\newpage

\noindent
\includegraphics[
    width=0.9\textwidth,
    trim=0 0 0 945,
    clip
]{figures/Objective-Drift-Case.pdf}

\captionof{figure}{
Case of objective drift. The Agent keeps optimizing a certain subtask but overlooks the end-to-end final delivery target. It fails to carry out subsequent steps, making the final output inconsistent with requirements.
}
\label{fig:Objective Drift}

\section{List of all professional software in Workflow-GYM }
\label{app:softwarelist}

The full list of all the softwares included in \bench{} tasks is shown at table \ref{tab:software_list}.

\begin{table}[t]
\centering
\small
\begin{tabular}{ll}
\toprule
3D Slicer & Aladin \\
Anki & AntConc \\
Audacity & Blender \\
CellProfiler & ClustalW \\
Darktable & DWSIM \\
FreeCAD & FreeMind \\
GeoDa & GeoGebra \\
GIMP & GNU Radio 4.0 \\
GnuCash & Godot \\
Grass GIS & HandBrake \\
HomeBank & Horizon EDA \\
JabRef & jamovi \\
JASP & Kdenlive \\
KiCad & KMyMoney \\
KNIME & LibreCAD \\
MeshLab & MITK Workbench \\
Mixxx & MoneyManager \\
MySQL & OBS Studio \\
OpenLCA & OpenModelica \\
OpenPLC Editor & OpenShot \\
Orange & ParaView \\
Photoshop & QGIS \\
Qucs-S & SAGA GIS \\
SatScan & Shotcut \\
SQL-Server & Sweet Home 3D \\
Tiled & VESTA \\
weka & Wireshark \\
CapCut & JLCEDA \\
\bottomrule
\end{tabular}
\caption{Professional software environments included in Workflow-GYM.}
\label{tab:software_list}
\end{table}